\documentclass{article}

\usepackage[final]{neurips_data_2024}

\usepackage[utf8]{inputenc} % allow utf-8 input
\usepackage[T1]{fontenc}    % use 8-bit T1 fonts
\usepackage{hyperref}       % hyperlinks
\usepackage{url}            % simple URL typesetting
\usepackage{booktabs}       % professional-quality tables
\usepackage{amsfonts}       % blackboard math symbols
\usepackage{nicefrac}       % compact symbols for 1/2, etc.
\usepackage{microtype}      % microtypography
\usepackage{xcolor}         % colors
\usepackage{xcolor}         % colors
\usepackage{ulem}           % strikeout text
\usepackage[pdftex]{graphicx}
\usepackage{wrapfig} % for wrapping tables
\usepackage{amsmath}
\usepackage{multirow}
\usepackage{gensymb} % This package provides the \degree command

\usepackage{authblk}

\usepackage[capitalize]{cleveref}
\usepackage{colortbl}
\renewcommand{\ss}[1]{{\ifshowcomments\textcolor{purple}{SS: #1}\else \fi}}

\newif\ifshowcomments

\showcommentsfalse

\title{emg2pose: A Large and Diverse Benchmark for Surface Electromyographic Hand Pose Estimation}

\author{Sasha Salter\thanks{Equal contribution.} }
\author{Richard Warren$^*$}
\author{Collin Schlager$^*$}
\author{Adrian Spurr}
\author{Shangchen Han}
\author{Rohin Bhasin\thanks{Work done while at Meta.} }
\author{Yujun Cai}
\author{Peter Walkington}
\author{Anuoluwapo Bolarinwa}
\author{Robert Wang}
\author{Nathan Danielson}
\author{Josh Merel$^\dag$}
\author{Eftychios Pnevmatikakis}
\author{Jesse Marshall}
\affil{Reality Labs, Meta}

\begin{document}

\maketitle
\vspace{-1.33cm}
\begin{figure}[h!]
  % \centering
  \includegraphics[width=1\textwidth,]{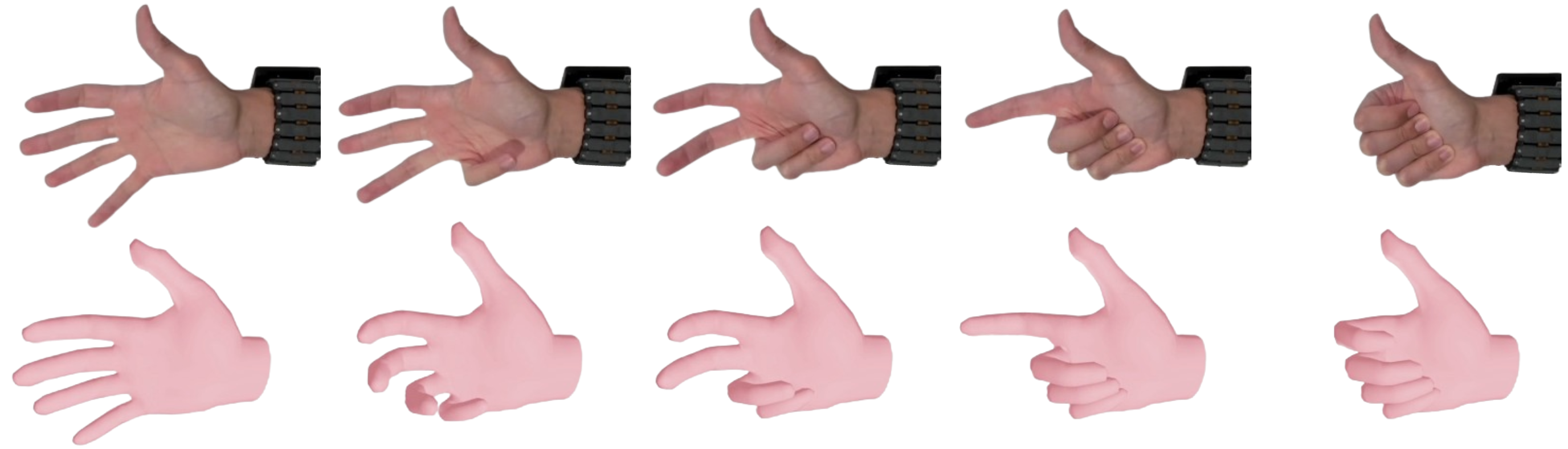}
  \caption{We introduce the \textit{emg2pose} dataset and benchmark to facilitate the development of pose estimation models from sEMG. Our \textit{vemg2pose} model is capable of estimating in real-time hand pose (lower) from held-out users wearing an sEMG wristband (top). See text for further details.}
  \label{fig: EMG2Pose online}
\end{figure}

\begin{abstract}
\begin{enumerate}
    Hands are the primary means through which humans interact with the world. Reliable and always-available hand pose inference could yield new and intuitive control schemes for human-computer interactions, particularly in virtual and augmented reality. 
Computer vision is effective but requires one or multiple cameras and can struggle with occlusions, limited field of view, and poor lighting. 
Wearable wrist-based surface electromyography (sEMG) presents a promising alternative as an always-available modality sensing muscle activities that drive hand motion. However, sEMG signals are strongly dependent on user anatomy and sensor placement, and existing sEMG models have required hundreds of users and device placements to effectively generalize. 
% % OLD
% To facilitate progress on sEMG pose inference, we introduce the \textit{emg2pose benchmark}, which is to our knowledge the first publicly available dataset of high-quality hand pose labels and wrist sEMG recordings. 
% NEW
To facilitate progress on sEMG pose inference, we introduce the \textit{emg2pose benchmark}, the largest publicly available dataset of high-quality hand pose labels and wrist sEMG recordings. 
\textit{emg2pose} contains $2$kHz, $16$ channel sEMG and pose labels from a 26-camera motion capture rig for $193$ users, $370$ hours, and
$29$ \textit{stages} with diverse gestures - a scale comparable to vision-based hand pose datasets. We provide competitive baselines and challenging tasks evaluating real-world generalization scenarios: \textit{held-out users, sensor placements}, and \textit{stages}. \textit{emg2pose} provides the machine learning community a platform for exploring complex generalization problems, holding potential to significantly enhance the development of sEMG-based human-computer interactions.

\end{enumerate}
\end{abstract}

\section{Introduction}
\label{sec: intro}

Despite rapid progress in computing hardware and software, current input devices can be inefficient and non-intuitive for new and emerging computing platforms.
This is particularly evident for spatial interactions, such as those encountered in virtual and augmented reality, where conventional input devices like controllers, keyboards, and mice do not always offer intuitive control schemes 
nor sufficient degrees of freedom 
to enable precise control (e.g., object manipulation).
Interactions based on hand movements offer a high-dimensional continuous input 
that is instinctive, universal, and particularly well suited to spatial interactions \citep{han2020megatrack}. Furthermore, existing input schemes can be viewed as low dimensional summaries 
of hand movements, for instance a mouse click tells you that a finger has pressed a button. 
As such, hand kinematics is a potentially holistic and encompassing modality, covering existing inputs and extending them in a natural manner.
High fidelity hand tracking enables various AR/VR applications including gaming \citep{han2020megatrack}, virtual teaching \citep{shrestha2022aimusicguru}, teleoperations \citep{santos2012increasing,darvish2023teleoperation}, haptics \citep{scheggi2015touch}, embodied realism \citep{wang2020rgb2hands}, sports analytics \citep{gatt2020accuracy}, and healthcare and rehabilitation \citep{krasoulis2017improved}.

Given the high utility and broad appeal of effective hand pose estimation, there have been diverse approaches developed across many sensing modalities: optical approaches (e.g. monocular, multi-view, depth-based, motion capture, infrared) using fixed  \citep{cai2018weakly,mueller2018ganerated,ge2016robust,supanvcivc2018depth, park20203d} or head-mounted cameras \citep{han2018online}; wearable data gloves using magnetic \citep{parizi2019auraring}, inertial \citep{yang2021hand}, capacitative \citep{truong2018capband}, and stretch sensors \citep{shen2016soft, tashakori2024capturing, luo2021learning}; smart rings \citep{parizi2019auraring}; wrist and forearm wearables that use impedance tomography \citep{zhang2015tomo}, inertial measurement units \citep{laput2019sensing}, acoustics \citep{ViBand2016} or ultrasound \citep{mcintosh2017echoflex}. Each modality comes with its own hardware constraints and limitations. Optical approaches can struggle with occlusions, poor lighting conditions, and limited field of view, and often require multiple cameras for effective inference, which places constraints on the overall size of the device. Alternatively, glove wearables can hinder dexterous manipulation \citep{roda2020effect} and forearm wearables typically only support discrete gesture classification. 

Surface electromyography (sEMG) sensing on the wrist or forearm provides an appealing alternative that does not struggle with occlusion, field of view, poor lighting, or physical encumberance. sEMG uses electrodes on the skin to measure electrical potentials generated by muscles during movement \citep{stashuk2001emg}. Specifically,
sEMG detects the electrical activity that occurs when spinal motor neurons activate the muscle fibers
that drive motion \citep{merletti2016surface}. As such, sEMG is particularly well suited for kinematic inference and numerous approaches have been developed \citep{liu2021neuropose,quivira2018translating,sosin2018continuous,simpetru2022accurate}. 
Nevertheless, learning a universal sEMG-to-pose model that \textit{generalizes} to new participants and kinematics is particularly challenging. This is due to sEMG sensing 
containing many axes of variation, primarily: \textit{user anatomy}, \textit{sensor placement}, and \textit{hand kinematics} \citep{ctrl2024generic,liu2021neuropose}. User anatomy and sensor placement both influence the locations of the sensors relative to the muscles. 
Hand kinematics influence what combination of muscle activities are sensed.
Given the number of generative dimensions, sEMG models are particularly data-hungry \citep{ctrl2024generic},
necessitating many samples across these axes to effectively learn universal models that generalize (see \cref{sec: dataset_scale_analysis} experiments). Existing datasets lack scale across each of these generative dimensions, thus hindering the development of generic models \citep{atzori2014electromyography}.
% Existing datasets are not open sourced and are relatively small (<20 participant) and brief (<20 minutes per participant), thus hindering the development of generic models \citep{liu2021neuropose, simpetru2022sensing}. 

Another complication of sEMG is that it encodes muscle activity, which relates more closely to motion than the pose that we would like to recover. As such, direct pose inference from sEMG is particularly challenging (see \cref{sec: res}), potentially requiring reasoning over long historical sEMG sequences to disambiguate pose from sequences of indirect motion measurements. Extracting relevant information from long sequences, or contexts, in the presence of ambiguity has been extensively explored in fields such as CV \citep{brunetti2018computer,kirillov2023segment,pan2018two}, natural language \citep{achiam2023gpt,kojima2022large,gu2021efficiently}, and robotics \citep{lauri2022partially,dunion2024conditional,jang2022bc}. Despite this, prior sEMG works have shown promising results for personalized or single-user pose inference settings \citep{liu2021neuropose,simpetru2022accurate}. 

To facilitate progress toward developing universal sEMG-to-pose models, we introduce the \textit{emg2pose benchmark} dataset, a large-scale dataset of simultaneously recorded high-fidelity wrist sEMG recordings and hand pose labels. High resolution sEMG recordings are obtained with the sEMG-RD wrist band 
\citep{ctrl2024generic}(see \cref{sec: sEMG background})
and high precision pose labels are obtained from a 26-camera motion capture rig that offers benefits compared to multi-view computer vision \citep{liu2021neuropose,sosin2018continuous}. To our knowledge, this is the 
% OLD
% only 
% NEW
largest
publicly-available sEMG hand pose dataset, spanning $193$ users, $370$ hours, and $29$ diverse kinematic categories, called \textit{stages}, each containing diverse low-level behaviors, called \textit{gestures}. In addition, the $80$M labelled frames that our dataset contains compares favourably with even the newest and largest CV equivalents \citep{sener2022assembly101, yu2020humbi} in both number of frames as well as subjects (see \cref{tab: cv datasets}). We additionally provide three competitive baselines and challenging hand pose inference benchmarks, investigating generalization to unseen \textit{users}, \textit{stages}, and \textit{user-stage} combinations. Instructions regarding accessing and using the \textit{emg2pose benchmark} is provided in \url{https://github.com/facebookresearch/emg2pose}. Given the high potential impact of sEMG input devices, and the similar research challenges to existing fields, we believe this benchmark will be of great value to the machine learning community. 
\section{Related Work}

\begin{wraptable}{r}{0.6\textwidth}
    \vspace{-7mm}
    \tiny
    % \begin{table} %[ht]
        % \footnotesize
        \caption{The largest publicly available sEMG datasets. We report the maximum number of gestures performed for \citet{du2017surface,atzori2014electromyography} and non amputee subjects for \citet{atzori2014electromyography}. \textit{\# Gest.} represents the number of poses participants were instructed to perform. Note that pose category definitions vary significantly across datasets. }
        \label{tab: semg datasets}
        \begin{tabular}{lccccccc}
            \toprule
            Dataset & \# Sess. & \# Subj.  & \# Sess. / subj & \# Gest. & Inc. Pose\\ 
            \midrule
            \citet{palermo2017repeatability} & 100 & 10  & 10 & 7 & No  \\
            % \hline
            \citet{amma2015advancing} & 25 & 5  & 5 & 27 & No\\
            % \hline
            \citet{du2017surface} & 69 & 23  & 3 & 22 & No\\
            % \hline
            \citet{jiang2021open} & 40 & 20 & 2 & 34 & No\\
            \citet{atzori2014electromyography} & 67 & 67 & 1 & 53 & Yes\\
            \midrule
            Ours & 751 & 193 & 4 & 50 & Yes\\
            \toprule
        \end{tabular}
        \vspace{-3mm}
    % \end{table}
\end{wraptable}

\textbf{sEMG Datasets:} There are several publicly available sEMG datasets for tasks other than pose regression, specifically pose (sequence) classification. Data have been collected with either clinical-grade high-density electrode arrays and amplifiers 
\citep{amma2015advancing,du2017surface,jiang2021open}
or consumer-grade hardware that has fewer channels and lower temporal resolution \citep{palermo2017repeatability}. Clinical-grade hardware offers hundreds of recording channels and acquisition rates >1 kHz but are impractical due to lengthy donning procedures that include shaving the skin before applying conductive gel and the electrode arrays. In contrast, existing consumer-grade hardware is easier to deploy, but is limited by lower bandwidth and channel counts and thus may not provide the level of fidelity required for pose estimation. In contrast, our dataset uses the sEMG-RD band \citep{ctrl2024generic}, that can be quickly donned,  records 16 channels at >2 kHz and has proven performant for generalized pose classification modelling. 

\citet{atzori2014electromyography} released a pose regression dataset on consumer-grade emg technologies. This dataset is composed of  $37$ hours of simultaneously recorded kinematics and sEMG, over a total of $67$ sensor placements. In contrast, our dataset includes $193$ users, $370$ hours, and $751$ sessions, which should allow us to train models that generalize favourably across these axes (see \cref{tab: semg datasets} for scale comparison with existing datasets). Our dataset contains \textit{gesture} categories as well as joint angles. 

\textbf{Pose Regression from sEMG:} Several papers have studied pose regression from sEMG, although without open sourcing datasets. \citet{liu2021neuropose} use the MyoBand to estimate hand pose across diverse movements in an 11 participant dataset. They test sEMG decoding models of hand pose across users and sessions with both convolutional (NeuroPose; see \cref{sec: baselines}) and LSTM architectures. \citet{simpetru2022sensing} (SensingDynamics; see \cref{sec: baselines}) use a clinic-grade system to collect several dozen minute datasets in a set of 13 participants. They use a custom 3D convolutional architecture to predict hand joint angles, landmark positions, and grip force, reporting tracking with low error in a held-out test set within each participant.
These datasets are limited in scale, with only 11 or 13 participants, and 15 or 20 minutes of data per participant for \citet{liu2021neuropose, simpetru2022sensing}, respectively, likely limiting generalization across users. In contrast, our dataset includes $193$ users and $370$ hours, aiding the development of generic models that generalize across users (see \cref{sec: dataset_scale_analysis}).

\textbf{Pose from Computer Vision:} Computer vision (CV) based hand pose estimation has received considerable attention in recent years, usually taking depth, RGB, or both as input, and leveraging large open-sourced datasets \citep{mueller2017real, mueller2018ganerated, spurr2018cross, spurr2020weakly, spurr2021self, wan2019self, boukhayma20193d}. Labels are either obtained using marker-based motion capture \citep{fan2023arctic} - whose markers create an input distributional shift due to lack of markers during deployment - or using alternate approaches with lower quality labels or inputs, such as multi-view cameras \citep{zimmermann2019freihand, moon2024dataset}, synthetic data \citep{zimmermann2017learning}, 
and magnetic sensors \citep{yuan2017bighand2}. In contrast, motion capture markers afford high quality labels for sEMG, but they do not affect the data from which predictions are generated.

\begin{wraptable}{r}{6.3cm}
    \vspace{-6mm}
    \scriptsize
    % \begin{table} %[ht]
        % \footnotesize
        \caption{Largest CV hand datasets. The \textit{per hand} row counts the data from the left and right hands independently, whereas the \textit{across hands} row pools those data.}
        \label{tab: cv datasets}
        \begin{tabular}{lccc}
            \toprule
            Dataset & \# Frames & \# Subjects & \# FPS \\
            \midrule
            \citet{yuan2017bighand2} & 2.2M & 10 & 60\\
            % \hline
            \citet{moon2020interhand2} & 2.6M & 27 & 5-30\\
            % \hline
            \citet{moon2024dataset} & 1.5M & 10 & 5-30 \\
            % \hline
            \citet{samarth2020contactpose} & 2.9M & 50 & n/a \\
            % \hline
            \citet{fan2023arctic} & 2.1M & 10 & 30 \\
            % \hline
            \citet{liu2022hoi4d} & 2.4M & 4 & n/a \\
            % \hline
            \citet{sener2022assembly101} & 111M & 53 & n/a \\
            % \hline
            \citet{yu2020humbi} & 24M & 453 & 60 \\
            \midrule
            Ours (per hand) & 80M & 193 & 60 \\
            Ours (across hands) & 40M & 193 & 60 \\
            \toprule
        \end{tabular}
        \vspace{-5mm}
    % \end{table}
\end{wraptable}

The gestural diversity of CV-based datasets mostly focuses on exploring the full static pose space of the hand \citep{yuan2017bighand2, zimmermann2017learning, zimmermann2019freihand}, interaction with objects \citep{fan2023arctic, samarth2020contactpose, hampali2020honnotate} or hand-hand interactions \citep{moon2020interhand2, moon2024dataset}. Conversely, our dataset focuses on movements of the hand because sEMG, unlike CV, is more closely related to motion than pose. Furthermore, our dataset has 80M frames and 193 subjects, comparing favorably to CV datasets (see \cref{tab: cv datasets}, reporting million frames, subjects and fps).

\textbf{Pose from Other Modalities:} 
In addition to vision and sEMG, there exists a diverse range of additional wearable approaches to pose inference (see \cref{sec: intro}), which typically focus on pose (sequence) classification. For example, \citet{achenbach2023give} released a dataset for pose classification using commercially available sensor gloves. Other datasets typically use bespoke hardware and are small in scale, with the exception of a large (50 participant, 25 class) dataset available for classification using commercially available smartwatches 
\citep{laput2019sensing}.
\section{emg2pose Benchmark}
\label{sec: emg2pose_benchmark}

\subsection{sEMG Device}
\label{sec: sEMG background}

Data are collected using the 16 channel bipolar sEMG-RD wrist band from \citet{ctrl2024generic}. They demonstrate the effectiveness of this device for generalized pose sequence classification across $6400$ participants, the largest study to date. This high performance is achieved without the need for high-density sEMG platforms \citep{amma2015advancing}, with a similar form factor and ease of use to other low-density platforms \citep{rawat2016evaluating} (see \cref{fig: EMG2Pose online,fig: data_collection} for a visual depiction of the device). 
In contrast to the previously used low-density Thalmic Labs Myo band \citep{liu2021neuropose} that streams data at $200$Hz, across $8$ channels and with $8$-bits, sEMG-RD senses at $2$kHz, across $16$ channels and with $12$-bits. 
For more details see \cref{sec: semg_sensing}.

\subsection{Dataset}
\label{sec: dataset}

\begin{figure}[h]
  \centering
  \includegraphics[width=1.\textwidth]{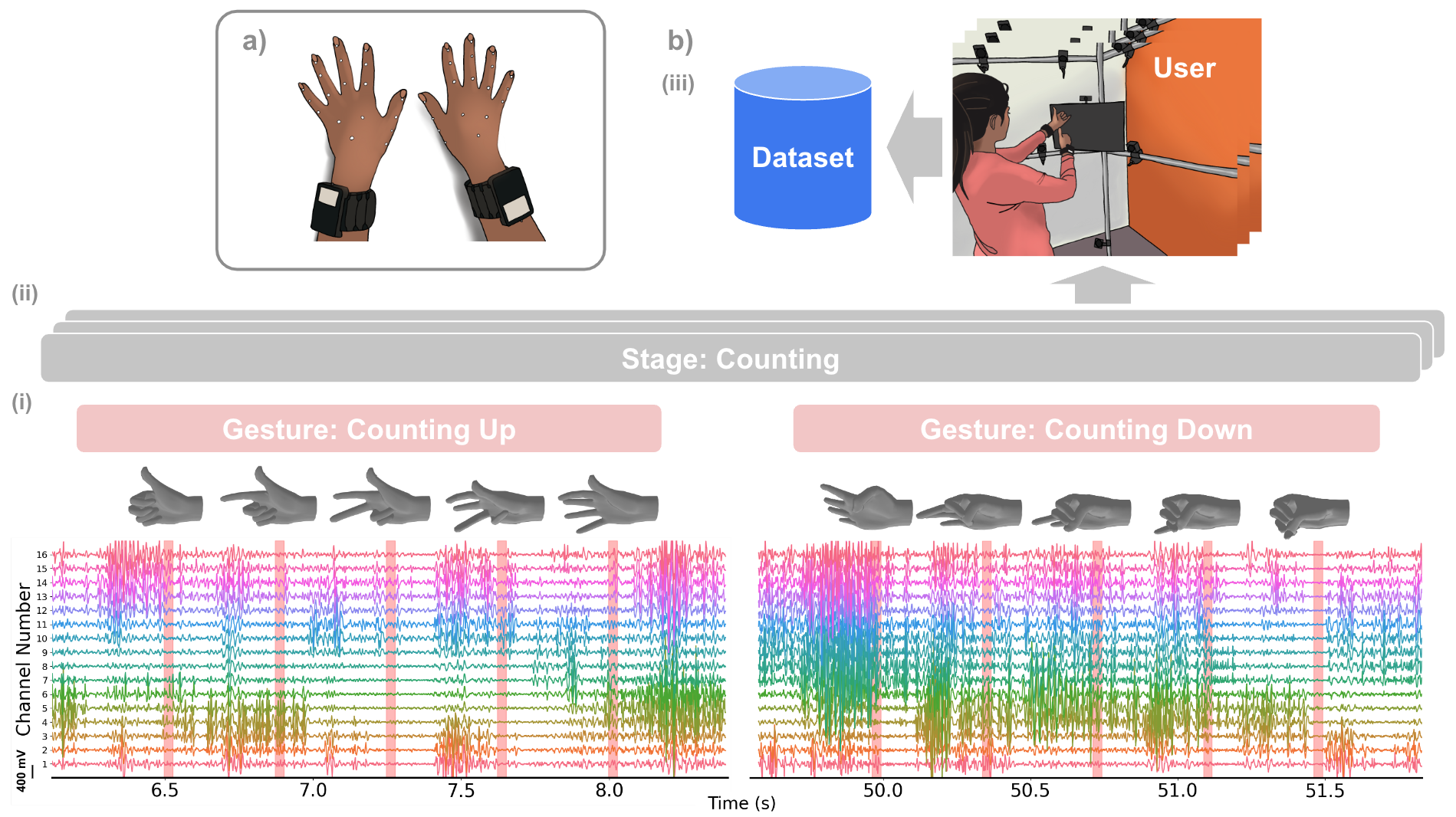}
  \caption{\textit{Dataset composition}: a) sEMG-RD wrist-band and motion capture marker (white dots) setup. b) Dataset breakdown. i) Users are prompted to perform a sequence of movement types (\textit{gestures}), such as counting up and down. sEMG and poses are recorded simultaneously. ii) Groups of specific gesture types comprise a \textit{stage}, such as counting. Stages are partitioned into train/val/test splits (see \cref{sec: held-out settings}).
  Our dataset consists of $29$ diverse stages. iii) Each of the $193$ users perform various stages, donning on-and-off the wrist band. In total we record $370$ hours of data.
  }
  \label{fig: data_collection}
  \vspace{-4mm}
\end{figure}

\begin{table}[ht]
    \scriptsize
    \centering
    \caption{\textit{emg2pose} dataset statistics, reporting mean and standard deviation. Three separate test sets measure generalization to new users, types of behaviors (stages), and user-behavior combinations (user, stage). Note that the overall hours is the sum of the hours across all splits. The number of hours counts the right-handed and left-handed data separately for each participant.}

    % The following is pasted from ctrl-jupyter
    \begin{tabular}{lccccccc}
    \hline
    & Train & \multicolumn{2}{|c|}{Val} & \multicolumn{3}{c|}{Test} & Overall \\
    &  & \multicolumn{1}{|c}{\textit{User}} & \multicolumn{1}{c|}{\textit{User, Stage}} & \textit{User} & \textit{Stage} & \multicolumn{1}{c|}{\textit{User, Stage}} &  \\
    \hline
    Subjects & 158 & 15 & 15 & 20 & 158 & 20 & 193 \\
    Unique stages & 23 & 23 & 6 & 23 & 6 & 6 & 29 \\
    Hours & 250.9 & 21.7 & 4.6 & 31.9 & 54.2 & 7.0 & 370.3 \\
    Hours / subject & 1.6 $\pm$ 0.4 & 1.4 $\pm$ 0.5 & 0.3 $\pm$ 0.1 & 1.6 $\pm$ 0.3 & 0.3 $\pm$ 0.1 & 0.3 $\pm$ 0.0 & 1.9 $\pm$ 0.5 \\
    Sessions / subject & 3.9 $\pm$ 0.6 & 3.8 $\pm$ 0.6 & 3.7 $\pm$ 0.6 & 3.9 $\pm$ 0.3 & 3.8 $\pm$ 0.7 & 3.8 $\pm$ 0.5 & 3.9 $\pm$ 0.6 \\
    \hline
    \end{tabular}
    
    \vspace{-1mm}
    \label{table:dataset_details}
\end{table}

Consenting participants (see \cref{sec: full datasheet}) stood in a $26$ camera motion capture array (\cref{sec: mocap}). A research assistant placed $19$ motion capture markers on each of the participants’ hands (\citet{han2018online}) and an sEMG-RD band on each wrist \citep{ctrl2024generic}. All sEMG and motion capture data were streamed to a real-time data acquisition system at $2$kHz and $60$ Hz, respectively. We time-aligned device streams using software timestamps, which we found to show less than $10$ms relative latency between devices. Motion capture data were post-processed using an offline inverse kinematics (IK) solver to reconstruct the joint angles of the hand (\cref{sec: full datasheet,sec: mocap}). The IK solver failed for $12.7\%$ of frames, typically due to simultaneously occluded markers. Finally, joints angles were linearly interpolated to 2~kHz to match the sample rate of sEMG.

Participants followed a standardized data collection protocol across a diverse set of 45-120~s \textit{stages} in which participants were prompted to perform either a mix of 3-5 similar \textit{gestures} in random orderings
(e.g. specific \textit{finger counting} orderings such as \textit{ascending} or \textit{descending}) or unconstrained freeform movements (see \cref{sec: stage_descriptions,sec: full datasheet} for further details). \textit{Stages} can be viewed as a categorization of \textit{gestures}. For example, the \textit{Counting} stage categorizes \textit{Counting Up} and \textit{Counting Down} gestures (see \cref{fig: data_collection}). 
During data collection, the majority of users donned on-and-off the device 4 times, with a small fraction only thrice.
Each group of stages with a single band placement is referred to as a \textit{session}. We report the prompted movements for each stage in detail in \cref{sec: stage_descriptions}.
During each stage, we prompted participants using videos and verbal instructions by the research assistant. Participants were instructed to move their hands across their body and between their waist and shoulders to ensure a range of different postures were sampled. See \cref{fig: data_collection} for a visualization of the data collection.

The full dataset is organized hierarchically by participant, session, and stage. In total, we collected data from $193$ participants, spanning $370$ hours, $751$ sessions, $29$ diverse stages (see \cref{sec: stage_descriptions} for further details and \cref{table:dataset_details} for statistics). Note that the number of hours counts the right-handed and left-handed data separately for each participant, although they were collected simultaneously. To our knowledge, this is the only open-sourced sEMG and motion capture dataset and is of similar scale to those in the CV literature \citep{yuan2017bighand2,brahmbhatt2020contactpose,moon2020interhand2,moon2024dataset}.
The entire dataset consists of $25,253$ HDF5 files, each consisting of time-aligned sEMG and joint angles for a single hand in a single stage.

\subsection{Tasks}
\label{sec: tasks}

The \textit{emg2pose benchmark} includes two benchmark tasks: \textit{pose regression} and \textit{pose tracking}.

\textbf{Regression:} For this task, previously explored in \citet{liu2021neuropose, simpetru2022accurate}, one must regress from sEMG to hand joint angle sequences. Without knowledge of the initial hand pose and velocity, this is a partially observable task \citep{spaan2012partially}, and thus particularly challenging for the reasons mentioned in \cref{sec: intro}. Pose regression is the most challenging task and is meant to promote continued research with applications including unimodal pose prediction in settings where computer vision is infeasible or unreliable.

\textbf{Tracking:} For this simpler task, one must regress from sEMG to hand joint angle sequences whilst being provided with the initial hand pose in the sequence. Providing the initial pose addresses the partial observability dilemma. Nevertheless, this task still poses the generalization challenges discussed in \cref{sec: intro}. The tracking task is meant to promote initial research and progress, and has several real-world applications. An effective tracker would provide great value in settings where: the user is prompted to match a given pose before tracking commences; visual pose prediction feedback is provided, allowing the user to adjust their pose to correct for erroneous initial predictions; and when ground truth pose estimates are intermittently available, such as from computer vision settings  whenever partial or full occlusions occur. 

\textbf{Evaluation:} We evaluate on $5$ second trajectories
and report test set mean absolute \textit{joint angular error} ($\degree$) and mean (Euclidean) \textit{landmark distance} (mm). Landmarks correspond to joint and fingertip Cartesians. We do not regress to wrist angles, which were not recorded for this dataset. Landmarks corresponding to the most proximal joint for fingers other than the thumb always have zero error because the wrist does not move. These landmarks are therefore excluded from our metrics. We obtain landmark locations by passing joint angles through a default hand model. This introduces bias, as it will not perfectly align with each user's anatomy. We leave addressing this limitation for future work. In real world applications, it will be important to not only improve mean performance for these metrics, but also lower percentile scores across the population.

\subsection{Held-Out Settings}
\label{sec: held-out settings}

Effective pose inference requires models that generalize across \textit{device placements}, \textit{users}, and \textit{hand kinematics}. Prior works have only investigated generalization across a subset of these axes, such as user \citep{liu2021neuropose, ctrl2024generic} or device placement \citep{liu2021neuropose, palermo2017repeatability}, but generalization to new types of kinematics has not been explicitly explored. In contrast, we provide three separate test sets intended to measure these axes independently. The statistics of each held-out scenario are reported in \cref{table:dataset_details}. In short, \textit{users} corresponds to unseen users, but in-distribution kinematics (\textit{stages}). \textit{Stages} represents unseen kinematic categories, but in-distribution users. Finally, \textit{users, stages} constitute held-out users and stages, and is of greatest value as the most encompassing real-world deployment setting. Both held out user scenarios all constitute new device placements, which vary across all sessions. We break down train, validation and test splits roughly using $0.7:0.1:0.2$ ratio with exact splits shown in \cref{table:dataset_details}. Held-out users are randomly sampled and held-out stages are chosen to be visually out-of-distribution with respect to the training stages. See \cref{fig: results stage breakdown} for a breakdown of which stages are in the training and held-out sets, \cref{table:stage_dataset_details} for details regarding each stage, and \cref{sec: data-collection-protocol} for further dataset details.

\subsection{Baselines}
\label{sec: baselines}

We provide three baselines: open-source re-implementations of the \textit{NeuroPose} and \textit{SensingDynamics} network architectures \citep{liu2021neuropose, simpetru2022sensing}, and a new \textit{vemg2pose} model. Algorithm details can be found in \cref{sec: algorithm_details}.

\textbf{vemg2pose:} sEMG meaures underlying muscle activity, and therefore relates more strongly to hand movements than the static pose of the hand. Therefore, \textit{vemg2pose} ("Velocity-based emg2pose") predicts joint angular velocities, which are then integrated to produce joint angle predictions. sEMG is first embedded via a causal strided convolutional \textit{featurizer}, which temporally down-samples sEMG from 2~kHz to 50~Hz. A \textit{Time-Depth Separable Convolution} (TDS) network is used for the featurizer, as it has been shown to be effective and parameter-efficient in the automatic speech recognition literature \citep{hannun2019sequence} (see \cref{sec: algorithm_details} for implementation details). The features at each time-step are then concatenated to the joint angle predictions at the previous time step and fed to an LSTM \textit{decoder}, which produces the next velocity prediction. Those velocities are added to the previous joint angles to produce the next prediction. vemg2pose is therefore auto-regressive with respect to its own predictions. Finally, predictions are linearly up-sampled to match the sample rate of the joint angles targets. For the tracking task, the initial joint angles are set to the ground truth, according to the motion capture labels. For the regression task, the initial state is also predicted by the decoder (see \cref{sec: algorithm_details} for further details).

\textbf{NeuroPose:} 
NeuroPose and vemg2pose differ in their prediction spaces and network architectures. Whereas vemg2pose predicts angular velocities, NeuroPose predicts joint angles directly. NeuroPose uses a U-Net architecture with residual bottleneck layers. Briefly, a convolutional encoder spatially and temporally down-samples sEMG while extracting features which are then refined via a stack of residual blocks. Finally, a decoder generates pose predictions at the original sample rate via convolutions and up-sampling layers. Because our sEMG device measures at 10x the temporal frequency and 2x the spatial frequency of the MyoBand used in \citet{liu2021neuropose}, we increase the temporal and spatial down and up-sampling of NeuroPose's featurizer and decoder (by 8x and 2x, respectively), such that the receptive field remains comparable to the original model. See \citet{liu2021neuropose} for full model details and \cref{sec: algorithm_details} for further details.

\textbf{SensingDynamics:} SensingDynamics and NeuroPose primarily differ in their architectures. Instead of a U-Net, SensingDynamics' featurizer comprises of 2d convolutions over sEMG channels and time, with learnable SMU activations \citep{biswas2021smu}, batch normalisation, circular padding across channels, and dropout layers. The decoder comprises of a 3-layered MLP. Uniquely, SensingDynamics additionally passes 20Hz low-passed filtered sEMG as input to the featurizer. See \citet{simpetru2022sensing} for full model details and \cref{sec: algorithm_details} for further details.   

\textbf{Training Setup:} All algorithms are trained to minimize the L1 error between predicted and ground truth joint angles as well as the Euclidean error between between predicted and ground truth fingertip locations.
The joint angle loss term has a weight of $1$ and the fingertip loss term has a weight of $.01$. We train on 1-6 seconds of non-overlapping trajectories. The training trajectory length - in addition to other hyperparameters - is optimized independently for each algorithm (see \cref{table:hyperparameters}). We train for 500 epochs with a 50 epoch early stopping criterion. Time-points for which motion capture data are not available are skipped during training and evaluation. We use a batch size of $64$ per GPU. We train on Amazon EC2 \texttt{g5.48xlarge} instances which have $8$x NVIDIA T4 GPUs for less than a day.

\section{Experiments}
\label{sec: res}

\subsection{Benchmark Results}

\begin{table}[ht]
    \small
    \centering
    \caption{\textit{Regression} test set results. Mean and standard deviation are reported across users. Bold indicates the significance of a Wilcoxon signed-rank test comparing vemg2pose to NeuroPose and Sensing Dynamics for each metric and condition (.01 threshold adjusted to .0008 via Bonferroni correction, see \cref{sec: statistical_analysis} for details).}.
    \renewcommand{\arraystretch}{1.2}  % Add more space between rows
    
    % The following is pasted from ctrl-jupyter
    \begin{tabular}{lcccccc}
    \toprule
    Test Set & Baseline & Angular Error ($\degree$) & Landmark Distance ($mm$) \\
    \midrule
    \multirow[t]{2}{*}{User} & SensingDynamics & 15.5 $\pm$ 1.4 & 21.8 $\pm$ 2.1 \\ & NeuroPose & 13.2 $\pm$ 1.1 & 17.5 $\pm$ 1.3 \\
     & vemg2pose & \textbf{12.2 $\pm$ 1.3} & \textbf{15.8 $\pm$ 1.9} \\
    \cline{1-4}
    \multirow[t]{2}{*}{Stage} & SensingDynamics & 18.8 $\pm$ 1.6 & 26.6 $\pm$ 2.0 \\ & NeuroPose & 17.2 $\pm$ 1.7 & 24.0 $\pm$ 2.1 \\
     & vemg2pose & \textbf{15.2 $\pm$ 1.6} & \textbf{20.4 $\pm$ 2.2} \\
    \cline{1-4}
    \multirow[t]{2}{*}{User, Stage} & SensingDynamics & 18.7 $\pm$ 1.6 & 27.2 $\pm$ 2.0 \\ & NeuroPose & 17.5 $\pm$ 1.5 & 24.9 $\pm$ 1.7 \\
     & vemg2pose & \textbf{15.8 $\pm$ 1.4} & \textbf{21.6 $\pm$ 2.0} \\
    \cline{1-4}
    \end{tabular}

    \vspace{-1mm}
    \label{table:regression_results}
\end{table}

\begin{figure}[h]
  \vspace{-3mm}
  \centering
  \includegraphics[width=1\textwidth]{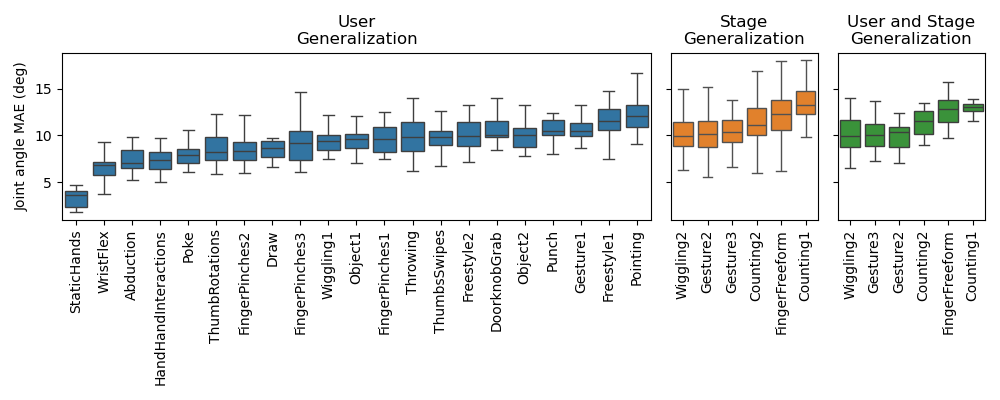}
  \vspace{-8mm}
  \caption{\textit{vemg2pose tracking performance break down} by stage and generalization condition. Distributions are over users. Note the variability in performance across stages. Each box shows the median and interquartile range (IQR), and whiskers show the minimum and maximum values that are within 1.5 times the IQR of the lower and upper quartiles.}
  \label{fig: results stage breakdown}
\end{figure}

We report \textit{regression} results in \cref{table:regression_results} and \textit{tracking} results in \cref{table:tracking_results}. We do not report standard deviation across model seeds, as we observed these to be negligible.
Results are further broken down by stage, finger, and joint in \cref{fig: results stage breakdown,fig: finger performance,fig: joint performance}, respectively. For the regression task, vemg2pose outperforms both NeuroPose and SensingDynamics with respect to both angular errors and landmark distances. In general, accuracy degrades most for the held-out \textit{user, stage} combination, which is the hardest of all transfer scenarios. For the tracking task - in which the initial ground truth pose is provided - errors are lower overall, as expected (see \cref{sec: tasks}). For this task, we do not report scores for NeuroPose and SensingDynamics, as these models were not originally designed to leverage knowledge of the initial ground truth pose during inference.

\begin{table}[ht]
    \small
    \centering
    \caption{\textit{Tracking} test set results. Mean and standard deviation are reported across users.}
    \renewcommand{\arraystretch}{1.2}  % Add more space between rows
    
    % The following is pasted from ctrl-jupyter
    \begin{tabular}{lccc}
    \toprule
     Test Set &  Baseline & Angular Error ($\degree$) & Landmark Distance ($mm$) \\
    \midrule
    User & vemg2pose & 7.7 $\pm$ 1.0 & 10.3 $\pm$ 1.5 \\
    \cline{1-4}
    Stage & vemg2pose & 11.2 $\pm$ 1.4 & 15.2 $\pm$ 1.9 \\
    \cline{1-4}
    User, Stage & vemg2pose & 11.0 $\pm$ 1.0 & 15.4 $\pm$ 1.4 \\
    \cline{1-4}
    \end{tabular}
    
    \vspace{-1mm}
    \label{table:tracking_results}
\end{table}

% NEW
Performance varies considerably across users for all models and tasks, potentially due to anatomical differences (\cref{table:regression_results,table:tracking_results}). Performance varies significantly across stages (\cref{fig: results stage breakdown}), which is likely a result of the amount and type of movements in each stage. Stages with limited movement (StaticHands, WristFlex) may be easier for the model track because they involve very limited postural transitions. Stages with complex hand poses and dynamic articulation of individual fingers (Gesture2, Pointing) are more challenging and have higher errors. Moreover, \cref{fig: finger performance} shows that performance varies significantly across fingers and finger joints, with the thumb the most reliably predicted, followed by the index, middle, ring, and pinky fingers. We also find that proximal joint angles of the fingers are easier to track than distal joint angles (\cref{fig: joint performance}). Together, this suggests that stages with high amounts of thumb movements (e.g. ThumbRotations) may be easier to track than those with more general finger movements (e.g. Freestyle1). 

\subsection{Analysis on Challenging Stages for Vision-Based Systems}

\begin{figure}[h!]
  % \vspace{-1mm}
  \centering
  \includegraphics[width=0.7\textwidth]{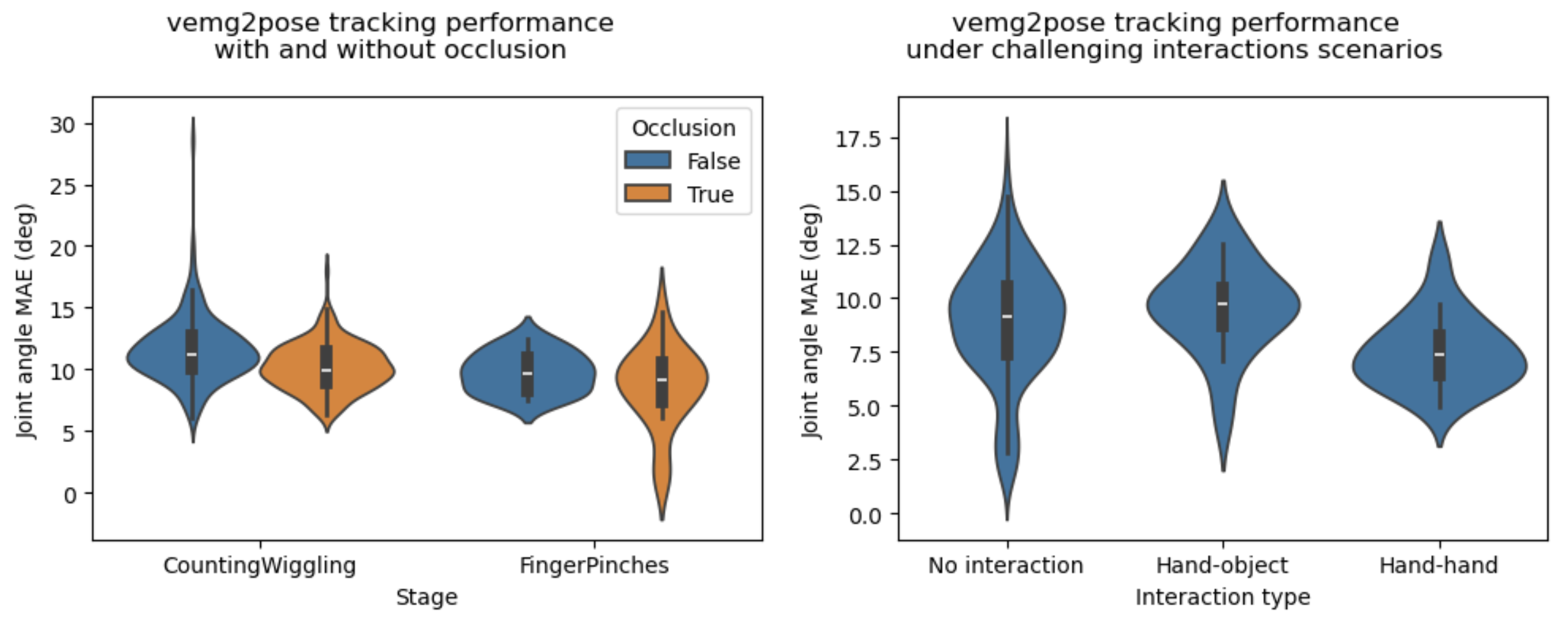}
  \vspace{-3mm}
  \caption{vemg2pose tracking results \textit{with/without occlusion (left) and physical interactions (right)}. Distributions are over users. See \cref{sec: challenging_stages_supp} for more details.
  }
  \label{fig: challenging_stages}
\end{figure}

Some stages were specifically designed to test behaviors that are known to be challenging for vision-based hand pose estimation (see \cref{sec: challenging_stages_supp} for details). We found that stages with hand-hand interactions or hand-object interactions have similar model performance compared to stages without such interactions (\cref{fig: challenging_stages}, right), although differences in behavioral distribution across these stages makes direct comparison challenging. Furthermore, we find that visual occlusion does not impact sEMG based pose reconstruction, as expected. Stages in which the hand is occluded from a CV based headset tracking system have similar accuracy compared to stages without occlusion in which the same behaviors are performed (\cref{fig: challenging_stages}, left).

\subsection{Qualitative Analysis}

\begin{figure}[h]
  \includegraphics[width=1\textwidth]{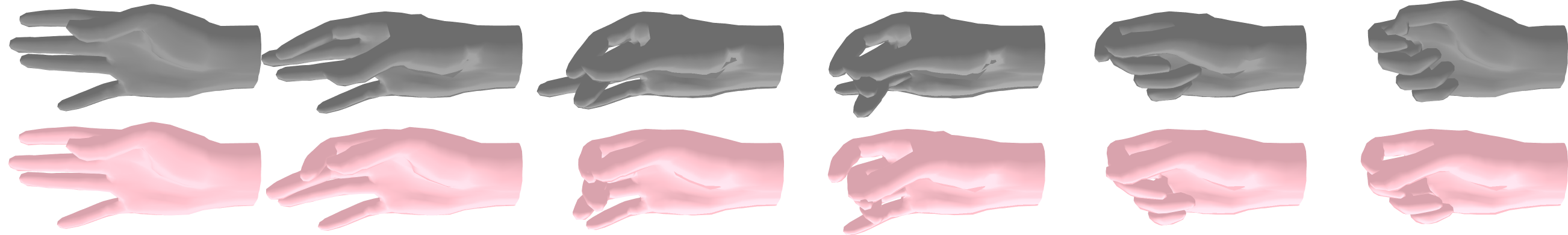}
  \caption{\textit{Median percentile held-out user and stage} (Counting2). Top: motion capture; bottom: vemg2pose, tracking predictions. Clips unroll evenly left-to-right over a $2$ second segment.}
  \label{fig: online rollout}
\end{figure}

We plot \textit{vemg2pose, tracking} real-time online and offline kinematic predictions for \textit{held-out users and stages} in \cref{fig: online rollout,fig: EMG2Pose online} (see \cref{sec: online_vemg2pose} for online setup details). This is the most challenging scenario, representing generalization to held-out kinematics, user anatomy, and device placement. For \cref{fig: online rollout}, we plot a median-performance representative held-out stage (Counting2, see \cref{table:stage_dataset_details}) and user. 
As seen, individual finger movements are mostly tracked, but not always.
We visualize top and bottom percentile (15\% and 85\%) offline kinematics for the held-out users and stages generalization setting in \cref{fig: v1,fig: v2,fig: v3,fig: v4}. 
In general, we observed three challenges specific to sEMG pose inference: angular drift due to sensing that strongly relates to pose derivatives (see the ring finger in \cref{fig: v4}); movements related to harder-to-sense intrinsic hand muscles, such as the finger adduction/abduction present in the "vulcan" gesture (\cref{table:stage_dataset_details}); movements related to smaller and fewer muscles, such as pinky (see \cref{fig: finger performance}) and distal joint motion (see \cref{fig: EMG2Pose online}).

\subsection{Dataset Scale Analysis}
\label{sec: dataset_scale_analysis}

\begin{figure}[h]
\vspace{-5mm}
  \centering
  \includegraphics[width=0.36\textwidth]{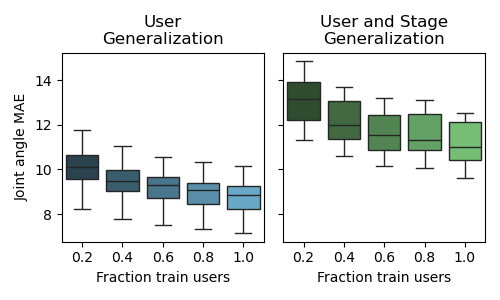}
  \includegraphics[width=0.58\textwidth]{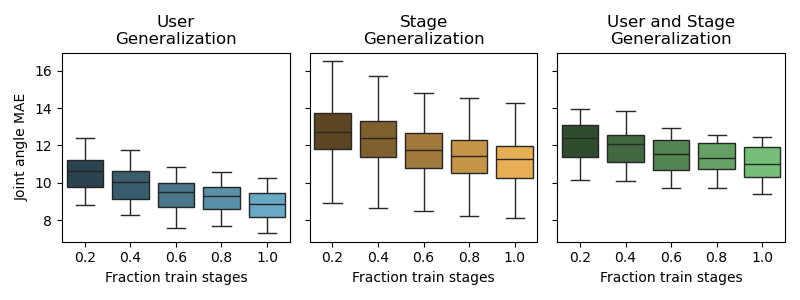}
  \caption{\textit{Generalization vs. number of training users (left two) or stages (right three) for vemg2pose tracking}. We subsampled the training users/stages but evaluated on the same held-out users/stages. As seen, performance improves with the number of training users/stages, demonstrating the importance of our dataset scale for effective generalization. Box plots take the same format as \cref{fig: results stage breakdown}.}
  \label{fig: training user scale}
\end{figure}

We ran experiments to \textit{demonstrate the importance of the scale} of our dataset for effective generalization. In \cref{fig: training user scale}, we show that increasing the number of training users considerably reduces the error for held-out users, perhaps because models are exposed to sEMG from users with a variety of wrist anatomies. We also show that increasing the number of stages per-user improves performance across all modes of generalization, demonstrating the importance of behavioural diversity.

\subsection{Quantifying Generalization Difficulty across Users and Stages}
\label{sec: dist-gap-gen}

\begin{figure}[h]
  \centering
  \includegraphics[width=0.48\textwidth]{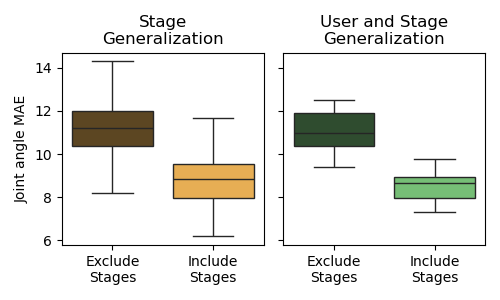}
  \includegraphics[width=0.48\textwidth]{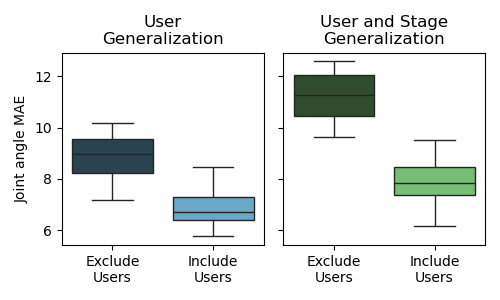}
  \caption{Excluding stages (left) or users (right) from the training set \textit{markedly decreases performance for these stages/users}. For the \textit{include stages/users} condition, we include $70$\% of the data from the held-out stages/users in the training set. For the \textit{exclude stages/users} condition we exclude that $70$\% entirely. Both test sets are identical allowing us to isolate the influence of holding out stages/users. Data are from a tracking task with a vemg2pose model. Distributions are over users.}
  \label{fig: held-out stage degredation}
\end{figure}

To directly quantify the \textit{difficulty of generalizing} across held-out \textit{stages} and \textit{users}, we performed experiments in which a subset of the data from the held-out users and stages were either folded into the training set or excluded entirely. \cref{fig: held-out stage degredation} shows that excluding specific users and stages from the training set markedly degrades performance, demonstrating the difficulty of generalizing across these dimensions. Refer to \cref{fig: held-out stage degredation} for a detailed description of the experimental setup.

\vspace{-4mm}
\section{Limitations and Future Work}
\label{sec: discussion}
\vspace{-1mm}

\textbf{Modelling: } We provide an initial investigation into generalized sEMG-to-pose modelling and open-source our baselines to the community. Nevertheless, there remains a plethora of unexplored, potentially fruitful sequence modelling directions, such as state space and diffusion-based methods. Pose estimation in the presence of uncertainty introduced by sensor noise and anatomical variability could also be addressed with probabilistic methods \citep{danelljan2020probabilistic}. Model personalization has also been shown to be beneficial \citep{ctrl2024generic,liu2021neuropose}, yet we do not explore this avenue here. In addition, our models obtain mean landmark distance errors that are higher than reported in the CV literature \citep{boukhayma20193d,mueller2017real}, despite having the advantage of not having to infer the wrist position or user's anatomy. Addressing this performance gap will be of great importance.
Finally, the lack of broader access to the sEMG-RD wrist-band \citep{ctrl2024generic} might be limiting, as this precludes human-in-the-loop testing of models. 

\textbf{Metrics: } Our landmark distance metrics use a default hand model to convert joint angles to joint positions. The mismatch between the hand model and user anatomy will bias this metric. In general, our metrics do not capture the physical plausibility of model predictions. For example, we have observed that vemg2pose sometimes predicts unfeasible kinematics, such as intra-finger penetration. Providing metrics that capture these failure modes will be of value, especially for embodied applications \citep{yuan2023physdiff}. Simulators of the hand \citep{caggiano2022myosuite} could be leveraged in a manner similar to \citet{yuan2023physdiff} to ensure physical constraints are adhered to. Finally, our held-out \textit{user, stage} test scenario is meant to best represent real-world in the wild performance. Nevertheless, it does not cover a potential range of signal aggressors such as: electrode-skin contact artifacts; impedance changes from sweat; electrical interference from external devices; and non-stationarity due to muscle fatigue. While these aggressors likely play a minor role in sEMG variability, they may be important to include in future datasets and test sets. 

\textbf{Dataset: } We discuss dataset limitations in \cref{sec: dataset limitations}.

\textbf{Ethical and Societal Implications: } We discuss ethical and societal implications in \cref{sec: societal implications}.
\vspace{-2mm}
\section{Conclusion}

We introduce the \textit{emg2pose benchmark}, 
% OLD
% the first large,
% NEW
the largest, 
diverse, and open-source dataset of high-fidelity sEMG recordings and hand pose labels. We introduce competitive benchmark models that can track or regress to hand pose for held-out users, stages and sessions, although there remains significant room to improve these models in future research. Due to the myriad sources of variability in sEMG signals, deciphering the relationship between sEMG and movement in a manner that generalizes across people and kinematics will likely require new algorithmic advances, taking inspiration from related machine learning fields. Large datasets like \textit{emg2pose} should thus facilitate progress in both sEMG decoding and machine learning applied to biosignals more broadly. Progress will enable intuitive,  high-dimensional human-computer interfaces that we perceive as extensions of ourselves.

\section{Acknowledgements}
We thank Patrick Kaifosh and TR Reardon for their sponsorship and vision and the entire CTRL-labs team for their collaboration and support. We thank Carl Hewitt and Migmar Tsering for help with data collection, Steve Olsen and Mark Hogan for assistance setting up motion capture recordings, John Choi and Diogo Peixoto for technical assistance and advice, and Dano Morrison and Sunaina Rajani for assistance with visualizations.

\bibliographystyle{abbrvnat}
% \bibliography{references.bib}

\begin{thebibliography}{78}
  \providecommand{\natexlab}[1]{#1}
  \providecommand{\url}[1]{\texttt{#1}}
  \expandafter\ifx\csname urlstyle\endcsname\relax
    \providecommand{\doi}[1]{doi: #1}\else
    \providecommand{\doi}{doi: \begingroup \urlstyle{rm}\Url}\fi

  \bibitem[Achenbach et~al.(2023)Achenbach, Laux, Purdack, M{\"u}ller, and
    G{\"o}bel]{achenbach2023give}
  P.~Achenbach, S.~Laux, D.~Purdack, P.~N. M{\"u}ller, and S.~G{\"o}bel.
  \newblock Give me a sign: Using data gloves for static hand-shape recognition.
  \newblock \emph{Sensors}, 23\penalty0 (24):\penalty0 9847, 2023.

  \bibitem[Achiam et~al.(2023)Achiam, Adler, Agarwal, Ahmad, Akkaya, Aleman,
    Almeida, Altenschmidt, Altman, Anadkat, et~al.]{achiam2023gpt}
  J.~Achiam, S.~Adler, S.~Agarwal, L.~Ahmad, I.~Akkaya, F.~L. Aleman, D.~Almeida,
    J.~Altenschmidt, S.~Altman, S.~Anadkat, et~al.
  \newblock Gpt-4 technical report.
  \newblock \emph{arXiv preprint arXiv:2303.08774}, 2023.

  \bibitem[Amma et~al.(2015)Amma, Krings, B{\"o}er, and
    Schultz]{amma2015advancing}
  C.~Amma, T.~Krings, J.~B{\"o}er, and T.~Schultz.
  \newblock Advancing muscle-computer interfaces with high-density
    electromyography.
  \newblock In \emph{Proceedings of the 33rd Annual ACM Conference on Human
    Factors in Computing Systems}, pages 929--938, 2015.

  \bibitem[Atzori et~al.(2014)Atzori, Gijsberts, Castellini, Caputo, Hager,
    Elsig, Giatsidis, Bassetto, and M{\"u}ller]{atzori2014electromyography}
  M.~Atzori, A.~Gijsberts, C.~Castellini, B.~Caputo, A.-G.~M. Hager, S.~Elsig,
    G.~Giatsidis, F.~Bassetto, and H.~M{\"u}ller.
  \newblock Electromyography data for non-invasive naturally-controlled robotic
    hand prostheses.
  \newblock \emph{Scientific data}, 1\penalty0 (1):\penalty0 1--13, 2014.

  \bibitem[Biswas et~al.(2021)Biswas, Kumar, Banerjee, and Pandey]{biswas2021smu}
  K.~Biswas, S.~Kumar, S.~Banerjee, and A.~K. Pandey.
  \newblock Smu: smooth activation function for deep networks using smoothing
    maximum technique.
  \newblock \emph{arXiv preprint arXiv:2111.04682}, 2021.

  \bibitem[Boukhayma et~al.(2019)Boukhayma, Bem, and Torr]{boukhayma20193d}
  A.~Boukhayma, R.~d. Bem, and P.~H. Torr.
  \newblock 3d hand shape and pose from images in the wild.
  \newblock In \emph{Proceedings of the IEEE/CVF Conference on Computer Vision
    and Pattern Recognition}, pages 10843--10852, 2019.

  \bibitem[Brahmbhatt et~al.(2020)Brahmbhatt, Tang, Twigg, Kemp, and
    Hays]{brahmbhatt2020contactpose}
  S.~Brahmbhatt, C.~Tang, C.~D. Twigg, C.~C. Kemp, and J.~Hays.
  \newblock Contactpose: A dataset of grasps with object contact and hand pose.
  \newblock In \emph{Computer Vision--ECCV 2020: 16th European Conference,
    Glasgow, UK, August 23--28, 2020, Proceedings, Part XIII 16}, pages 361--378.
    Springer, 2020.

  \bibitem[Brunetti et~al.(2018)Brunetti, Buongiorno, Trotta, and
    Bevilacqua]{brunetti2018computer}
  A.~Brunetti, D.~Buongiorno, G.~F. Trotta, and V.~Bevilacqua.
  \newblock Computer vision and deep learning techniques for pedestrian detection
    and tracking: A survey.
  \newblock \emph{Neurocomputing}, 300:\penalty0 17--33, 2018.

  \bibitem[Caggiano et~al.(2022)Caggiano, Wang, Durandau, Sartori, and
    Kumar]{caggiano2022myosuite}
  V.~Caggiano, H.~Wang, G.~Durandau, M.~Sartori, and V.~Kumar.
  \newblock Myosuite--a contact-rich simulation suite for musculoskeletal motor
    control.
  \newblock \emph{arXiv preprint arXiv:2205.13600}, 2022.

  \bibitem[Cai et~al.(2018)Cai, Ge, Cai, and Yuan]{cai2018weakly}
  Y.~Cai, L.~Ge, J.~Cai, and J.~Yuan.
  \newblock Weakly-supervised 3d hand pose estimation from monocular rgb images.
  \newblock In \emph{Proceedings of the European conference on computer vision
    (ECCV)}, pages 666--682, 2018.

  \bibitem[{CTRL{-}labs at Reality Labs} et~al.(2024){CTRL{-}labs at Reality
    Labs}, Sussillo, Kaifosh, and Reardon]{ctrl2024generic}
  {CTRL{-}labs at Reality Labs}, D.~Sussillo, P.~Kaifosh, and T.~Reardon.
  \newblock A generic noninvasive neuromotor interface for human-computer
    interaction.
  \newblock \emph{bioRxiv}, 2024.
  \newblock \doi{10.1101/2024.02.23.581779}.
  \newblock URL
    \url{https://www.biorxiv.org/content/early/2024/02/28/2024.02.23.581779}.

  \bibitem[Danelljan et~al.(2020)Danelljan, Gool, and
    Timofte]{danelljan2020probabilistic}
  M.~Danelljan, L.~V. Gool, and R.~Timofte.
  \newblock Probabilistic regression for visual tracking.
  \newblock In \emph{Proceedings of the IEEE/CVF conference on computer vision
    and pattern recognition}, pages 7183--7192, 2020.

  \bibitem[Darvish et~al.(2023)Darvish, Penco, Ramos, Cisneros, Pratt, Yoshida,
    Ivaldi, and Pucci]{darvish2023teleoperation}
  K.~Darvish, L.~Penco, J.~Ramos, R.~Cisneros, J.~Pratt, E.~Yoshida, S.~Ivaldi,
    and D.~Pucci.
  \newblock Teleoperation of humanoid robots: A survey.
  \newblock \emph{IEEE Transactions on Robotics}, 39\penalty0 (3):\penalty0
    1706--1727, 2023.

  \bibitem[Du et~al.(2017)Du, Jin, Wei, Hu, and Geng]{du2017surface}
  Y.~Du, W.~Jin, W.~Wei, Y.~Hu, and W.~Geng.
  \newblock Surface emg-based inter-session gesture recognition enhanced by deep
    domain adaptation.
  \newblock \emph{Sensors}, 17\penalty0 (3):\penalty0 458, 2017.

  \bibitem[Dunion et~al.(2024)Dunion, McInroe, Luck, Hanna, and
    Albrecht]{dunion2024conditional}
  M.~Dunion, T.~McInroe, K.~S. Luck, J.~Hanna, and S.~Albrecht.
  \newblock Conditional mutual information for disentangled representations in
    reinforcement learning.
  \newblock \emph{Advances in Neural Information Processing Systems}, 36, 2024.

  \bibitem[Fan et~al.(2023)Fan, Taheri, Tzionas, Kocabas, Kaufmann, Black, and
    Hilliges]{fan2023arctic}
  Z.~Fan, O.~Taheri, D.~Tzionas, M.~Kocabas, M.~Kaufmann, M.~J. Black, and
    O.~Hilliges.
  \newblock Arctic: A dataset for dexterous bimanual hand-object manipulation.
  \newblock In \emph{Proceedings of the IEEE/CVF Conference on Computer Vision
    and Pattern Recognition}, pages 12943--12954, 2023.

  \bibitem[Gatt et~al.(2020)Gatt, Allen, and Wheat]{gatt2020accuracy}
  I.~T. Gatt, T.~Allen, and J.~Wheat.
  \newblock Accuracy and repeatability of wrist joint angles in boxing using an
    electromagnetic tracking system.
  \newblock \emph{Sports Engineering}, 23:\penalty0 1--10, 2020.

  \bibitem[Ge et~al.(2016)Ge, Liang, Yuan, and Thalmann]{ge2016robust}
  L.~Ge, H.~Liang, J.~Yuan, and D.~Thalmann.
  \newblock Robust 3d hand pose estimation in single depth images: from
    single-view cnn to multi-view cnns.
  \newblock In \emph{Proceedings of the IEEE conference on computer vision and
    pattern recognition}, pages 3593--3601, 2016.

  \bibitem[Gebru et~al.(2021)Gebru, Morgenstern, Vecchione, Vaughan, Wallach,
    Iii, and Crawford]{gebru2021datasheets}
  T.~Gebru, J.~Morgenstern, B.~Vecchione, J.~W. Vaughan, H.~Wallach, H.~D. Iii,
    and K.~Crawford.
  \newblock Datasheets for datasets.
  \newblock \emph{Communications of the ACM}, 64\penalty0 (12):\penalty0 86--92,
    2021.

  \bibitem[Gu et~al.(2021)Gu, Goel, and R{\'e}]{gu2021efficiently}
  A.~Gu, K.~Goel, and C.~R{\'e}.
  \newblock Efficiently modeling long sequences with structured state spaces.
  \newblock \emph{arXiv preprint arXiv:2111.00396}, 2021.

  \bibitem[Hampali et~al.(2020)Hampali, Rad, Oberweger, and
    Lepetit]{hampali2020honnotate}
  S.~Hampali, M.~Rad, M.~Oberweger, and V.~Lepetit.
  \newblock Honnotate: A method for 3d annotation of hand and object poses.
  \newblock In \emph{Proceedings of the IEEE/CVF conference on computer vision
    and pattern recognition}, pages 3196--3206, 2020.

  \bibitem[Han et~al.(2018)Han, Liu, Wang, Ye, Twigg, and Kin]{han2018online}
  S.~Han, B.~Liu, R.~Wang, Y.~Ye, C.~D. Twigg, and K.~Kin.
  \newblock Online optical marker-based hand tracking with deep labels.
  \newblock \emph{Acm transactions on graphics (tog)}, 37\penalty0 (4):\penalty0
    1--10, 2018.

  \bibitem[Han et~al.(2020)Han, Liu, Cabezas, Twigg, Zhang, Petkau, Yu, Tai,
    Akbay, Wang, et~al.]{han2020megatrack}
  S.~Han, B.~Liu, R.~Cabezas, C.~D. Twigg, P.~Zhang, J.~Petkau, T.-H. Yu, C.-J.
    Tai, M.~Akbay, Z.~Wang, et~al.
  \newblock Megatrack: monochrome egocentric articulated hand-tracking for
    virtual reality.
  \newblock \emph{ACM Transactions on Graphics (ToG)}, 39\penalty0 (4):\penalty0
    87--1, 2020.

  \bibitem[Han et~al.(2022)Han, Wu, Zhang, Liu, Zhang, Wang, Si, Zhang, Cai,
    Hodan, Cabezas, Tran, Akbay, Yu, Keskin, and Wang]{han2022umetrack}
  S.~Han, P.~Wu, Y.~Zhang, B.~Liu, L.~Zhang, Z.~Wang, W.~Si, P.~Zhang, Y.~Cai,
    T.~Hodan, R.~Cabezas, L.~Tran, M.~Akbay, T.~Yu, C.~Keskin, and R.~Wang.
  \newblock Umetrack: Unified multi-view end-to-end hand tracking for {VR}.
  \newblock In \emph{{SIGGRAPH} Asia 2022 Conference Papers, {SA} 2022, Daegu,
    Republic of Korea, December 6-9, 2022}, 2022.

  \bibitem[Hannun et~al.(2019)Hannun, Lee, Xu, and Collobert]{hannun2019sequence}
  A.~Hannun, A.~Lee, Q.~Xu, and R.~Collobert.
  \newblock Sequence-to-sequence speech recognition with time-depth separable
    convolutions.
  \newblock \emph{arXiv preprint arXiv:1904.02619}, 2019.

  \bibitem[Ingram et~al.(2008)Ingram, K{\"o}rding, Howard, and
    Wolpert]{ingram2008statistics}
  J.~N. Ingram, K.~P. K{\"o}rding, I.~S. Howard, and D.~M. Wolpert.
  \newblock The statistics of natural hand movements.
  \newblock \emph{Experimental brain research}, 188:\penalty0 223--236, 2008.

  \bibitem[Jang et~al.(2022)Jang, Irpan, Khansari, Kappler, Ebert, Lynch, Levine,
    and Finn]{jang2022bc}
  E.~Jang, A.~Irpan, M.~Khansari, D.~Kappler, F.~Ebert, C.~Lynch, S.~Levine, and
    C.~Finn.
  \newblock Bc-z: Zero-shot task generalization with robotic imitation learning.
  \newblock In \emph{Conference on Robot Learning}, pages 991--1002. PMLR, 2022.

  \bibitem[Jiang et~al.(2021)Jiang, Liu, Fan, Ye, Dai, Clancy, Akay, and
    Chen]{jiang2021open}
  X.~Jiang, X.~Liu, J.~Fan, X.~Ye, C.~Dai, E.~A. Clancy, M.~Akay, and W.~Chen.
  \newblock Open access dataset, toolbox and benchmark processing results of
    high-density surface electromyogram recordings.
  \newblock \emph{IEEE Transactions on Neural Systems and Rehabilitation
    Engineering}, 29:\penalty0 1035--1046, 2021.

  \bibitem[Kenton and Toutanova(2019)]{kenton2019bert}
  J.~D. M.-W.~C. Kenton and L.~K. Toutanova.
  \newblock Bert: Pre-training of deep bidirectional transformers for language
    understanding.
  \newblock In \emph{Proceedings of naacL-HLT}, volume~1, page~2. Minneapolis,
    Minnesota, 2019.

  \bibitem[Kirillov et~al.(2023)Kirillov, Mintun, Ravi, Mao, Rolland, Gustafson,
    Xiao, Whitehead, Berg, Lo, et~al.]{kirillov2023segment}
  A.~Kirillov, E.~Mintun, N.~Ravi, H.~Mao, C.~Rolland, L.~Gustafson, T.~Xiao,
    S.~Whitehead, A.~C. Berg, W.-Y. Lo, et~al.
  \newblock Segment anything.
  \newblock In \emph{Proceedings of the IEEE/CVF International Conference on
    Computer Vision}, pages 4015--4026, 2023.

  \bibitem[Kojima et~al.(2022)Kojima, Gu, Reid, Matsuo, and
    Iwasawa]{kojima2022large}
  T.~Kojima, S.~S. Gu, M.~Reid, Y.~Matsuo, and Y.~Iwasawa.
  \newblock Large language models are zero-shot reasoners.
  \newblock \emph{Advances in neural information processing systems},
    35:\penalty0 22199--22213, 2022.

  \bibitem[Krasoulis et~al.(2017)Krasoulis, Kyranou, Erden, Nazarpour, and
    Vijayakumar]{krasoulis2017improved}
  A.~Krasoulis, I.~Kyranou, M.~S. Erden, K.~Nazarpour, and S.~Vijayakumar.
  \newblock Improved prosthetic hand control with concurrent use of myoelectric
    and inertial measurements.
  \newblock \emph{Journal of neuroengineering and rehabilitation}, 14:\penalty0
    1--14, 2017.

  \bibitem[Laput and Harrison(2019)]{laput2019sensing}
  G.~Laput and C.~Harrison.
  \newblock Sensing fine-grained hand activity with smartwatches.
  \newblock In \emph{Proceedings of the 2019 CHI Conference on Human Factors in
    Computing Systems}, pages 1--13, 2019.

  \bibitem[Laput et~al.(2016)Laput, Xiao, and Harrison]{ViBand2016}
  G.~Laput, R.~Xiao, and C.~Harrison.
  \newblock Viband: High-fidelity bio-acoustic sensing using commodity smartwatch
    accelerometers.
  \newblock In \emph{Proceedings of the 29th Annual Symposium on User Interface
    Software and Technology}, UIST '16, page 321–333, New York, NY, USA, 2016.
    Association for Computing Machinery.
  \newblock ISBN 9781450341899.
  \newblock \doi{10.1145/2984511.2984582}.
  \newblock URL \url{https://doi.org/10.1145/2984511.2984582}.

  \bibitem[Lauri et~al.(2022)Lauri, Hsu, and Pajarinen]{lauri2022partially}
  M.~Lauri, D.~Hsu, and J.~Pajarinen.
  \newblock Partially observable markov decision processes in robotics: A survey.
  \newblock \emph{IEEE Transactions on Robotics}, 39\penalty0 (1):\penalty0
    21--40, 2022.

  \bibitem[Liu et~al.(2021)Liu, Zhang, and Gowda]{liu2021neuropose}
  Y.~Liu, S.~Zhang, and M.~Gowda.
  \newblock Neuropose: 3d hand pose tracking using emg wearables.
  \newblock In \emph{Proceedings of the Web Conference 2021}, pages 1471--1482,
    2021.

  \bibitem[Liu et~al.(2022)Liu, Liu, Jiang, Lyu, Wan, Shen, Liang, Fu, Wang, and
    Yi]{liu2022hoi4d}
  Y.~Liu, Y.~Liu, C.~Jiang, K.~Lyu, W.~Wan, H.~Shen, B.~Liang, Z.~Fu, H.~Wang,
    and L.~Yi.
  \newblock Hoi4d: A 4d egocentric dataset for category-level human-object
    interaction.
  \newblock In \emph{Proceedings of the IEEE/CVF Conference on Computer Vision
    and Pattern Recognition}, pages 21013--21022, 2022.

  \bibitem[Luo et~al.(2021)Luo, Li, Sharma, Shou, Wu, Foshey, Li, Palacios,
    Torralba, and Matusik]{luo2021learning}
  Y.~Luo, Y.~Li, P.~Sharma, W.~Shou, K.~Wu, M.~Foshey, B.~Li, T.~Palacios,
    A.~Torralba, and W.~Matusik.
  \newblock Learning human--environment interactions using conformal tactile
    textiles.
  \newblock \emph{Nature Electronics}, 4\penalty0 (3):\penalty0 193--201, 2021.

  \bibitem[McIntosh et~al.(2017)McIntosh, Marzo, Fraser, and
    Phillips]{mcintosh2017echoflex}
  J.~McIntosh, A.~Marzo, M.~Fraser, and C.~Phillips.
  \newblock Echoflex: Hand gesture recognition using ultrasound imaging.
  \newblock In \emph{Proceedings of the 2017 CHI Conference on Human Factors in
    Computing Systems}, pages 1923--1934, 2017.

  \bibitem[Merletti and Farina(2016)]{merletti2016surface}
  R.~Merletti and D.~Farina.
  \newblock \emph{Surface electromyography: physiology, engineering, and
    applications}.
  \newblock John Wiley \& Sons, 2016.

  \bibitem[Moon et~al.(2020)Moon, Yu, Wen, Shiratori, and
    Lee]{moon2020interhand2}
  G.~Moon, S.-I. Yu, H.~Wen, T.~Shiratori, and K.~M. Lee.
  \newblock Interhand2. 6m: A dataset and baseline for 3d interacting hand pose
    estimation from a single rgb image.
  \newblock In \emph{Computer Vision--ECCV 2020: 16th European Conference,
    Glasgow, UK, August 23--28, 2020, Proceedings, Part XX 16}, pages 548--564.
    Springer, 2020.

  \bibitem[Moon et~al.(2024)Moon, Saito, Xu, Joshi, Buffalini, Bellan, Rosen,
    Richardson, Mize, De~Bree, et~al.]{moon2024dataset}
  G.~Moon, S.~Saito, W.~Xu, R.~Joshi, J.~Buffalini, H.~Bellan, N.~Rosen,
    J.~Richardson, M.~Mize, P.~De~Bree, et~al.
  \newblock A dataset of relighted 3d interacting hands.
  \newblock \emph{Advances in Neural Information Processing Systems}, 36, 2024.

  \bibitem[Mueller et~al.(2017)Mueller, Mehta, Sotnychenko, Sridhar, Casas, and
    Theobalt]{mueller2017real}
  F.~Mueller, D.~Mehta, O.~Sotnychenko, S.~Sridhar, D.~Casas, and C.~Theobalt.
  \newblock Real-time hand tracking under occlusion from an egocentric rgb-d
    sensor.
  \newblock In \emph{Proceedings of the IEEE International Conference on Computer
    Vision}, pages 1154--1163, 2017.

  \bibitem[Mueller et~al.(2018)Mueller, Bernard, Sotnychenko, Mehta, Sridhar,
    Casas, and Theobalt]{mueller2018ganerated}
  F.~Mueller, F.~Bernard, O.~Sotnychenko, D.~Mehta, S.~Sridhar, D.~Casas, and
    C.~Theobalt.
  \newblock Ganerated hands for real-time 3d hand tracking from monocular rgb.
  \newblock In \emph{Proceedings of the IEEE conference on computer vision and
    pattern recognition}, pages 49--59, 2018.

  \bibitem[Palermo et~al.(2017)Palermo, Cognolato, Gijsberts, M{\"u}ller, Caputo,
    and Atzori]{palermo2017repeatability}
  F.~Palermo, M.~Cognolato, A.~Gijsberts, H.~M{\"u}ller, B.~Caputo, and
    M.~Atzori.
  \newblock Repeatability of grasp recognition for robotic hand prosthesis
    control based on semg data.
  \newblock In \emph{2017 International Conference on Rehabilitation Robotics
    (ICORR)}, pages 1154--1159. IEEE, 2017.

  \bibitem[Pan et~al.(2018)Pan, Luo, Shi, and Tang]{pan2018two}
  X.~Pan, P.~Luo, J.~Shi, and X.~Tang.
  \newblock Two at once: Enhancing learning and generalization capacities via
    ibn-net.
  \newblock In \emph{Proceedings of the european conference on computer vision
    (ECCV)}, pages 464--479, 2018.

  \bibitem[Parizi et~al.(2019)Parizi, Whitmire, and Patel]{parizi2019auraring}
  F.~S. Parizi, E.~Whitmire, and S.~Patel.
  \newblock Auraring: Precise electromagnetic finger tracking.
  \newblock \emph{Proceedings of the ACM on Interactive, Mobile, Wearable and
    Ubiquitous Technologies}, 3\penalty0 (4):\penalty0 1--28, 2019.

  \bibitem[Park et~al.(2020)Park, Kim, and Woo]{park20203d}
  G.~Park, T.-K. Kim, and W.~Woo.
  \newblock 3d hand pose estimation with a single infrared camera via domain
    transfer learning.
  \newblock In \emph{2020 IEEE International Symposium on Mixed and Augmented
    Reality (ISMAR)}, pages 588--599. IEEE, 2020.

  \bibitem[{Plotly Technologies Inc.}(2015)]{plotly}
  {Plotly Technologies Inc.}
  \newblock Collaborative data science, 2015.
  \newblock URL \url{https://plot.ly}.

  \bibitem[Quivira et~al.(2018)Quivira, Koike-Akino, Wang, and
    Erdogmus]{quivira2018translating}
  F.~Quivira, T.~Koike-Akino, Y.~Wang, and D.~Erdogmus.
  \newblock Translating semg signals to continuous hand poses using recurrent
    neural networks.
  \newblock In \emph{2018 IEEE EMBS International Conference on Biomedical \&
    Health Informatics (BHI)}, pages 166--169. IEEE, 2018.

  \bibitem[Rawat et~al.(2016)Rawat, Vats, and Kumar]{rawat2016evaluating}
  S.~Rawat, S.~Vats, and P.~Kumar.
  \newblock Evaluating and exploring the myo armband.
  \newblock In \emph{2016 International Conference System Modeling \& Advancement
    in Research Trends (SMART)}, pages 115--120. IEEE, 2016.

  \bibitem[Roda-Sales et~al.(2020)Roda-Sales, Sancho-Bru, Vergara,
    Gracia-Ib{\'a}{\~n}ez, and Jarque-Bou]{roda2020effect}
  A.~Roda-Sales, J.~L. Sancho-Bru, M.~Vergara, V.~Gracia-Ib{\'a}{\~n}ez, and
    N.~J. Jarque-Bou.
  \newblock Effect on manual skills of wearing instrumented gloves during
    manipulation.
  \newblock \emph{Journal of biomechanics}, 98:\penalty0 109512, 2020.

  \bibitem[Samarth et~al.(2020)Samarth, Chengcheng, Christopher, Charles, and
    James]{samarth2020contactpose}
  B.~Samarth, T.~Chengcheng, D.~T. Christopher, C.~K. Charles, and H.~James.
  \newblock Contactpose: A dataset of grasps with object contact and hand pose.
  \newblock In \emph{European Conference on Computer Vision (ECCV)}, 2020.

  \bibitem[Santos~Carreras(2012)]{santos2012increasing}
  L.~Santos~Carreras.
  \newblock Increasing haptic fidelity and ergonomics in teleoperated surgery.
  \newblock Technical report, EPFL, 2012.

  \bibitem[Scheggi et~al.(2015)Scheggi, Meli, Pacchierotti, and
    Prattichizzo]{scheggi2015touch}
  S.~Scheggi, L.~Meli, C.~Pacchierotti, and D.~Prattichizzo.
  \newblock Touch the virtual reality: using the leap motion controller for hand
    tracking and wearable tactile devices for immersive haptic rendering.
  \newblock In \emph{ACM SIGGRAPH 2015 Posters}, pages 1--1. 2015.

  \bibitem[Sener et~al.(2022)Sener, Chatterjee, Shelepov, He, Singhania, Wang,
    and Yao]{sener2022assembly101}
  F.~Sener, D.~Chatterjee, D.~Shelepov, K.~He, D.~Singhania, R.~Wang, and A.~Yao.
  \newblock Assembly101: A large-scale multi-view video dataset for understanding
    procedural activities.
  \newblock In \emph{Proceedings of the IEEE/CVF Conference on Computer Vision
    and Pattern Recognition}, pages 21096--21106, 2022.

  \bibitem[Shen et~al.(2016)Shen, Yi, Li, Lo, Chen, Hu, and Wang]{shen2016soft}
  Z.~Shen, J.~Yi, X.~Li, M.~H.~P. Lo, M.~Z. Chen, Y.~Hu, and Z.~Wang.
  \newblock A soft stretchable bending sensor and data glove applications.
  \newblock \emph{Robotics and biomimetics}, 3\penalty0 (1):\penalty0 22, 2016.

  \bibitem[Shrestha et~al.(2022)Shrestha, Ferm{\"u}ller, Huang, Win, Zukerman,
    Parameshwara, and Aloimonos]{shrestha2022aimusicguru}
  S.~Shrestha, C.~Ferm{\"u}ller, T.~Huang, P.~T. Win, A.~Zukerman, C.~M.
    Parameshwara, and Y.~Aloimonos.
  \newblock Aimusicguru: Music assisted human pose correction.
  \newblock \emph{arXiv preprint arXiv:2203.12829}, 2022.

  \bibitem[S{\^\i}mpetru et~al.(2022{\natexlab{a}})S{\^\i}mpetru, Arkudas, Braun,
    Osswald, de~Oliveira, Eskofier, Kinfe, and Del~Vecchio]{simpetru2022sensing}
  R.~C. S{\^\i}mpetru, A.~Arkudas, D.~I. Braun, M.~Osswald, D.~S. de~Oliveira,
    B.~Eskofier, T.~M. Kinfe, and A.~Del~Vecchio.
  \newblock Sensing the full dynamics of the human hand with a neural interface
    and deep learning.
  \newblock \emph{bioRxiv}, pages 2022--07, 2022{\natexlab{a}}.

  \bibitem[S{\^\i}mpetru et~al.(2022{\natexlab{b}})S{\^\i}mpetru, Osswald, Braun,
    Oliveira, Cakici, and Del~Vecchio]{simpetru2022accurate}
  R.~C. S{\^\i}mpetru, M.~Osswald, D.~I. Braun, D.~S. Oliveira, A.~L. Cakici, and
    A.~Del~Vecchio.
  \newblock Accurate continuous prediction of 14 degrees of freedom of the hand
    from myoelectrical signals through convolutive deep learning.
  \newblock In \emph{2022 44th Annual International Conference of the IEEE
    Engineering in Medicine \& Biology Society (EMBC)}, pages 702--706. IEEE,
    2022{\natexlab{b}}.

  \bibitem[Sosin et~al.(2018)Sosin, Kudenko, and Shpilman]{sosin2018continuous}
  I.~Sosin, D.~Kudenko, and A.~Shpilman.
  \newblock Continuous gesture recognition from semg sensor data with recurrent
    neural networks and adversarial domain adaptation.
  \newblock In \emph{2018 15Th international conference on control, automation,
    robotics and vision (ICARCV)}, pages 1436--1441. IEEE, 2018.

  \bibitem[Spaan(2012)]{spaan2012partially}
  M.~T. Spaan.
  \newblock Partially observable markov decision processes.
  \newblock In \emph{Reinforcement learning: State-of-the-art}, pages 387--414.
    Springer, 2012.

  \bibitem[Spurr et~al.(2018)Spurr, Song, Park, and Hilliges]{spurr2018cross}
  A.~Spurr, J.~Song, S.~Park, and O.~Hilliges.
  \newblock Cross-modal deep variational hand pose estimation.
  \newblock In \emph{Proceedings of the IEEE conference on computer vision and
    pattern recognition}, pages 89--98, 2018.

  \bibitem[Spurr et~al.(2020)Spurr, Iqbal, Molchanov, Hilliges, and
    Kautz]{spurr2020weakly}
  A.~Spurr, U.~Iqbal, P.~Molchanov, O.~Hilliges, and J.~Kautz.
  \newblock Weakly supervised 3d hand pose estimation via biomechanical
    constraints.
  \newblock In \emph{European conference on computer vision}, pages 211--228.
    Springer, 2020.

  \bibitem[Spurr et~al.(2021)Spurr, Dahiya, Wang, Zhang, and
    Hilliges]{spurr2021self}
  A.~Spurr, A.~Dahiya, X.~Wang, X.~Zhang, and O.~Hilliges.
  \newblock Self-supervised 3d hand pose estimation from monocular rgb via
    contrastive learning.
  \newblock In \emph{Proceedings of the IEEE/CVF international conference on
    computer vision}, pages 11230--11239, 2021.

  \bibitem[Stashuk(2001)]{stashuk2001emg}
  D.~Stashuk.
  \newblock Emg signal decomposition: how can it be accomplished and used?
  \newblock \emph{Journal of Electromyography and Kinesiology}, 11\penalty0
    (3):\penalty0 151--173, 2001.

  \bibitem[Supan{\v{c}}i{\v{c}} et~al.(2018)Supan{\v{c}}i{\v{c}}, Rogez, Yang,
    Shotton, and Ramanan]{supanvcivc2018depth}
  J.~S. Supan{\v{c}}i{\v{c}}, G.~Rogez, Y.~Yang, J.~Shotton, and D.~Ramanan.
  \newblock Depth-based hand pose estimation: methods, data, and challenges.
  \newblock \emph{International Journal of Computer Vision}, 126:\penalty0
    1180--1198, 2018.

  \bibitem[Tashakori et~al.(2024)Tashakori, Jiang, Servati, Soltanian, Narayana,
    Le, Nakayama, Yang, Wang, Eng, et~al.]{tashakori2024capturing}
  A.~Tashakori, Z.~Jiang, A.~Servati, S.~Soltanian, H.~Narayana, K.~Le,
    C.~Nakayama, C.-l. Yang, Z.~J. Wang, J.~J. Eng, et~al.
  \newblock Capturing complex hand movements and object interactions using
    machine learning-powered stretchable smart textile gloves.
  \newblock \emph{Nature Machine Intelligence}, 6\penalty0 (1):\penalty0
    106--118, 2024.

  \bibitem[Truong et~al.(2018)Truong, Zhang, Muncuk, Nguyen, Bui, Nguyen, Lv,
    Chowdhury, Dinh, and Vu]{truong2018capband}
  H.~Truong, S.~Zhang, U.~Muncuk, P.~Nguyen, N.~Bui, A.~Nguyen, Q.~Lv,
    K.~Chowdhury, T.~Dinh, and T.~Vu.
  \newblock Capband: Battery-free successive capacitance sensing wristband for
    hand gesture recognition.
  \newblock In \emph{Proceedings of the 16th ACM Conference on Embedded Networked
    Sensor Systems}, pages 54--67, 2018.

  \bibitem[Wan et~al.(2019)Wan, Probst, Gool, and Yao]{wan2019self}
  C.~Wan, T.~Probst, L.~V. Gool, and A.~Yao.
  \newblock Self-supervised 3d hand pose estimation through training by fitting.
  \newblock In \emph{Proceedings of the IEEE/CVF conference on computer vision
    and pattern recognition}, pages 10853--10862, 2019.

  \bibitem[Wang et~al.(2020)Wang, Mueller, Bernard, Sorli, Sotnychenko, Qian,
    Otaduy, Casas, and Theobalt]{wang2020rgb2hands}
  J.~Wang, F.~Mueller, F.~Bernard, S.~Sorli, O.~Sotnychenko, N.~Qian, M.~A.
    Otaduy, D.~Casas, and C.~Theobalt.
  \newblock Rgb2hands: real-time tracking of 3d hand interactions from monocular
    rgb video.
  \newblock \emph{ACM Transactions on Graphics (ToG)}, 39\penalty0 (6):\penalty0
    1--16, 2020.

  \bibitem[Yang et~al.(2021)Yang, Yan, van Beijnum, Li, and
    Veltink]{yang2021hand}
  Z.~Yang, S.~Yan, B.-J.~F. van Beijnum, B.~Li, and P.~H. Veltink.
  \newblock Hand-finger pose estimation using inertial sensors, magnetic sensors
    and a magnet.
  \newblock \emph{IEEE sensors journal}, 21\penalty0 (16):\penalty0 18115--18122,
    2021.

  \bibitem[Yu et~al.(2020)Yu, Yoon, Lee, Venkatesh, Park, Yu, and
    Park]{yu2020humbi}
  Z.~Yu, J.~S. Yoon, I.~K. Lee, P.~Venkatesh, J.~Park, J.~Yu, and H.~S. Park.
  \newblock Humbi: A large multiview dataset of human body expressions.
  \newblock In \emph{Proceedings of the IEEE/CVF Conference on Computer Vision
    and Pattern Recognition}, pages 2990--3000, 2020.

  \bibitem[Yuan et~al.(2017)Yuan, Ye, Stenger, Jain, and Kim]{yuan2017bighand2}
  S.~Yuan, Q.~Ye, B.~Stenger, S.~Jain, and T.-K. Kim.
  \newblock Bighand2. 2m benchmark: Hand pose dataset and state of the art
    analysis.
  \newblock In \emph{Proceedings of the IEEE conference on computer vision and
    pattern recognition}, pages 4866--4874, 2017.

  \bibitem[Yuan et~al.(2023)Yuan, Song, Iqbal, Vahdat, and
    Kautz]{yuan2023physdiff}
  Y.~Yuan, J.~Song, U.~Iqbal, A.~Vahdat, and J.~Kautz.
  \newblock Physdiff: Physics-guided human motion diffusion model.
  \newblock In \emph{Proceedings of the IEEE/CVF International Conference on
    Computer Vision}, pages 16010--16021, 2023.

  \bibitem[Zhang and Harrison(2015)]{zhang2015tomo}
  Y.~Zhang and C.~Harrison.
  \newblock Tomo: Wearable, low-cost electrical impedance tomography for hand
    gesture recognition.
  \newblock In \emph{Proceedings of the 28th Annual ACM Symposium on User
    Interface Software \& Technology}, pages 167--173, 2015.

  \bibitem[Zimmermann and Brox(2017)]{zimmermann2017learning}
  C.~Zimmermann and T.~Brox.
  \newblock Learning to estimate 3d hand pose from single rgb images.
  \newblock In \emph{Proceedings of the IEEE international conference on computer
    vision}, pages 4903--4911, 2017.

  \bibitem[Zimmermann et~al.(2019)Zimmermann, Ceylan, Yang, Russell, Argus, and
    Brox]{zimmermann2019freihand}
  C.~Zimmermann, D.~Ceylan, J.~Yang, B.~Russell, M.~Argus, and T.~Brox.
  \newblock Freihand: A dataset for markerless capture of hand pose and shape
    from single rgb images.
  \newblock In \emph{Proceedings of the IEEE/CVF International Conference on
    Computer Vision}, pages 813--822, 2019.

  \end{thebibliography}

\newpage
\section*{Checklist}
\label{sec: checklist}

\begin{enumerate}

\item For all authors...
\begin{enumerate}
  \item Do the main claims made in the abstract and introduction accurately reflect the paper's contributions and scope?
    \answerYes{}
  \item Did you describe the limitations of your work?
    \answerYes{} See \cref{sec: discussion}.
  \item Did you discuss any potential negative societal impacts of your work?
    \answerYes{} See \cref{sec: societal implications}.
  \item Have you read the ethics review guidelines and ensured that your paper conforms to them?
    \answerYes{}
\end{enumerate}

\item If you are including theoretical results...
\begin{enumerate}
  \item Did you state the full set of assumptions of all theoretical results?
    \answerNA{}
	\item Did you include complete proofs of all theoretical results?
    \answerNA{}
\end{enumerate}

\item If you ran experiments (e.g. for benchmarks)...
\begin{enumerate}
  \item Did you include the code, data, and instructions needed to reproduce the main experimental results (either in the supplemental material or as a URL)?
    \answerYes{} In \cref{sec: intro} we link to \url{https://github.com/facebookresearch/emg2pose} that provides all the relevant information.
  \item Did you specify all the training details (e.g., data splits, hyperparameters, how they were chosen)?
    \answerYes{} See \cref{sec: emg2pose_benchmark,sec: full datasheet,sec: dataset_app_details,sec: algorithm_details,table:dataset_details,table:stage_dataset_details,table:hyperparameters}.
	\item Did you report error bars (e.g., with respect to the random seed after running experiments multiple times)?
    \answerYes{} For example, see \cref{table:regression_results,table:tracking_results,fig: results stage breakdown}. We include standard deviation and percentile statistics calculated across users, stages. We do not report errors bars due to multiple model random seeds, as we saw very minimal differences.
	\item Did you include the total amount of compute and the type of resources used (e.g., type of GPUs, internal cluster, or cloud provider)?
    \answerYes{} See \cref{sec: baselines}.
\end{enumerate}

\item If you are using existing assets (e.g., code, data, models) or curating/releasing new assets...
\begin{enumerate}
  \item If your work uses existing assets, did you cite the creators?
    \answerYes{} See \cref{sec: baselines} for details on existing baselines that we provide and architectures we use. See \cref{sec: dataset,sec: mocap} for details on the Optitrack system used to collect data. See \cref{sec: semg_sensing} for wrist-band details. See \cref{sec: full datasheet} for details including python packages we build on.
  \item Did you mention the license of the assets?
    \answerYes{Dataset and code are licensed under a CC-BY-NC-SA 4.0 license, which is listed in \cref{sec: full datasheet} and repository.} 
  \item Did you include any new assets either in the supplemental material or as a URL?
    \answerYes{} We provide videos in zip format to reviewers and further instruction regarding how to access the dataset in \cref{sec: dataset_instructions}.
  \item Did you discuss whether and how consent was obtained from people whose data you're using/curating?
    \answerYes{} See \cref{sec: full datasheet}.
  \item Did you discuss whether the data you are using/curating contains personally identifiable information or offensive content?
    \answerYes{} In \cref{sec: dataset,sec: full datasheet} we discuss how we remove personally identifiable information.
\end{enumerate}

\item If you used crowdsourcing or conducted research with human subjects...
\begin{enumerate}
  \item Did you include the full text of instructions given to participants and screenshots, if applicable?
    \answerNo{We provide a detailed description of the experimental instructions and a list of the movements participants were asked to perform (see \cref{sec: full datasheet,sec: dataset_app_details}).} 
  \item Did you describe any potential participant risks, with links to Institutional Review Board (IRB) approvals, if applicable?
    \answerYes{We discuss the consenting process in \cref{sec: full datasheet} and the approval of all research under an external IRB} 
  \item Did you include the estimated hourly wage paid to participants and the total amount spent on participant compensation?
    \answerNA{We recruited all participants through a third-party vendor that determined their compensation via market rates. We give details in  \cref{sec: full datasheet}.} 
\end{enumerate}

\end{enumerate}

\appendix

% \newpage

\section{Datasheet}
\label{sec: full datasheet}

We provide a datasheet in accordance with \citet{gebru2021datasheets}.

\textbf{Motivation: } The motivation for \textit{emg2pose} is to address the lack of wide-spread, sufficiently large, non-invasive surface electromyographic (sEMG) datasets with high-quality ground-truth annotations for a concrete task. sEMG as a technology has the potential to revolutionize how humans interact with computers, and this public dataset is motivated to facilitate progress in this domain without needing specialized hardware. The task we consider is hand pose inference, as a potentially holistic and encompassing modality, with many biomimetic applications. This dataset was created by the CTRL-Labs research group within Reality Labs, Meta.

\textbf{Composition: } The entire dataset consists of $25,253$ HDF5 files, each consisting of time-aligned sEMG and joint angles for a single hand in a single stage. In total, we collected data from $193$ participants, spanning $370$ hours and $29$ diverse stages. The number of hours includes both the right-handed and left-handed data for each participant, which were collected simultaneously. Each HDF5 file includes sEMG data from one hand, the stage label, and the joint angles. sEMG is recorded at $2$kHz, high pass filtered at 40 Hz, and rescaled such that the noise floor has a standard deviation of 1. We also flip the sign of the left-handed EMG data to account for the reversal of polarity caused by wearing the band on the left vs. right hand. Additionally, the dataset includes a metadata file in CSV format containing dataset split information (train, val, and test). All metadata have been de-identified to remove any personally identifiable information and does not identify any sub-population. See \cref{sec: emg2pose_benchmark} for additional details on the dataset and \cref{table:dataset_details} for statistics about the dataset such as the number of participants, total duration, number of sessions and stages. See \cref{table:stage_dataset_details} for further details with regards to the stage composition. The configuration for the precise data splits used in our experiments can be found in the following link: \url{https://github.com/facebookresearch/emg2pose}.

\textbf{Collection Process: } We recruited participants through a third-party vendor, who compensated participants at market rates. All recruitment and on-boarding followed an external IRB-approved protocol. We provided participants with information about the study, and before study initiation asked them to review and sign an IRB-reviewed consent form.  We gave all participants the opportunity to ask questions before the study and were able to discontinue participation at any point. To ensure participant well-being, on-site research administrators monitored participants during the study protocol. All data have been de-identified to remove any personally identifiable metadata. Participants stood in a 26 camera motion capture array (\cref{sec: mocap}). A research assistant placed $19$ motion capture markers on each of the participant’s hands (\citet{han2018online}) and an sEMG-RD band on each wrist \citep{ctrl2024generic}(\cref{sec: mocap}). All sEMG and motion capture data were streamed to a real-time data acquisition system at $2$kHz and $60$ Hz, respectively (\cref{sec: semg_sensing,sec: mocap}). Participants followed a standardized data collection protocol across a diverse set of 30-120 s \textit{stages} in which participants were prompted to perform a mix of 3-5 gestures. We organized the data collection into two repetitions of two different groups of 15 and 26 stages with a different band placement for each. \ss{Left sentence may need to be modified} Each group of stages with a single band placement is referred as a \textit{session}. For further stage and data collection details see \cref{sec: data-collection-protocol,sec: stage_descriptions}. 

\textbf{Preprocessing/Cleaning/Labeling: } sEMG recordings in the dataset are sampled at 2~kHz with a bit depth of 12 bits, with a maximum signal amplitude of 6.6~mV, and are bandpass filtered with -3~dB cutoffs at 20~Hz and 850~Hz before digitization (see \cref{sec: semg_sensing}). 

Joint angles were estimated from motion capture recordings using a custom inverse kinematics pipeline using a personalized hand model according to \citet{han2018online}. Briefly, 19 reflective markers were attached to each hand, and their 3D coordinates were tracked via a commercial Optitrack system with 26 cameras around the participant. A ConvNet then assigned labels to each marker. The labeled markers were registered to positions on a calibrated hand mesh to determine landmark positions. An inverse kinematics solver produced the final joint angles. We applied a conservative 15 Hz low pass filter (\citet{ingram2008statistics}) to the final joint angles to ensure there is no residual jitter. The mean absolute difference between the filtered and unfiltered signal was only 0.32 degrees across 500 recordings.

This model produced an estimate of joint angles for the MCP, PIP and DIP joints for each finger as well as the IP, MCP and CMC joints of the thumb. Each joint had a degree of freedom for flexion and extension, while each MCP joint had an additional degree of freedom for abduction and adduction. Following joint angle estimation, we used a forward kinematic algorithm using a generic hand model to produce estimates of landmark positions \citep{han2022umetrack}. We used the center of each joint, as well as the fingertips, as landmark positions for evaluation. Finally, joint angles were low-pass filtered at 15~Hz to remove tracking noise, and temporally upsampled to match to 2~kHz sample rate of sEMG.

\textbf{Uses: } The dataset and the associated tooling are meant to be used only to advance sEMG-based research topics of interest within the academic community for purely non-commercial purposes and applications. Our code for baseline models, built on top of frameworks such as PyTorch, PyTorch Lightning and Hydra, is designed such that it can be easily extended to the exploration of different models and novel techniques for this task. The dataset and the associated code are not intended to be used in conjunction with any other data types.

\textbf{Distribution and Maintenance: } The dataset and the code to reproduce the baselines are accessible via \url{https://github.com/facebookresearch/emg2pose}. The dataset is hosted on Amazon S3 and the code to reproduce the baseline experiments on GitHub  under the CC-BY-NC-SA 4.0 license. We welcome contributions from the research community. Any future update, as well as ongoing maintenance such as tracking and resolving issues identified by the broader community, will be performed and distributed through the GitHub repository.

\section{Dataset Details}
\label{sec: dataset_app_details}

\begin{figure}[h]
  \centering
  \includegraphics[width=1\textwidth]{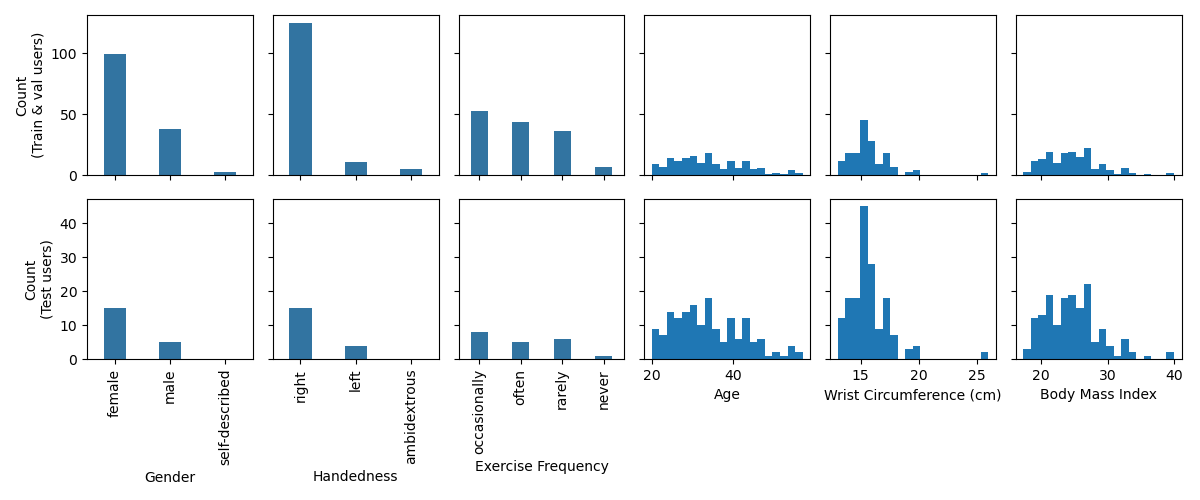}
  \caption{\textit{Participant demographic information.} Train and val users are shown in the top row, and test users on the bottom. Notice that test users are representative of the population of train/val users.}
  \label{fig: demographics}
\end{figure}

\subsection{sEMG Sensing}
\label{sec: semg_sensing}

sEMG data were collected using the sEMG-RD \citep{ctrl2024generic} consisting of $16$ differential electrode pairs utilizing dry gold-plated electrodes.
The 16 electrodes are arranged on a rigid ribbon, leaving a gap between electrodes 0 and 15 on the ulnar side of the wrist close to the ulnar styloid. Identical bands are worn on the left and right hands, with the same electrode indices aligning with the same anatomical features, but the polarity of the differential sensing being reversed. The band is tightened with an elastic strap and the size of the gap depends on the subject's wrist size and tightness. The band is manufactured in three different sizes to account for large changes in wrist size and the electrodes themselves are spring-loaded to further adjust across small variations in wrist sizes. In contrast, the previously used low-density Thalmic Labs Myo band \citep{rawat2016evaluating} only streams data at $200$Hz, across $8$ channels and at $8$-bits.

\subsection{Motion Capture}
\label{sec: mocap}

All motion capture data were collected using a 26 camera motion camera array at 60 Hz (Prime13W Optitrack) in an external data collection facility. Before data collection participants donned an sEMG band on each wrist and 19 3mm facial motion capture markers in order. We placed markers at the base of each fingernail and between the DIP and PIP and PIP and MCP joints of each finger. For the thumb we placed markers between the IP and MCP joint, on the MCP joint, and between the MCP and CMC joints. We additionally placed markers in a triangular pattern on the dorsal side of the hand \citep{han2018online}. Participants additionally wore a 3D printed frame of an XR headset, tethered to a PC, that was not relevant to the present data collection. Before collection, participants were asked to perform a series of 17 calibration gestures with each hand. These gestures were used as input to a custom optimization software that estimated the size of the hand and the position of motion capture markers relative to joints \citep{han2018online}. We saved personalized hand model information to a separate file to be used offline to estimate joint angles from the collected motion capture data. 

 During data collection, sEMG data were streamed over Bluetooth to a real-time data collection application. Motion capture data were recorded over ethernet using Motive2 (Naturalpoint) and then streamed to the same data collection pipeline. sEMG and motion capture datastreams were assigned software timestamps based on their arrival at the data pipeline. Internal testing bounded the relative latency between the two recording pathways to below 10 ms, approximately the Nyquist limit of the 60 Hz Optitrack recording. 

\subsubsection{Data Collection Protocol}
\label{sec: data-collection-protocol}
Data collection was divided into 4 different sessions (band placements). Participants performed two repetitions of two different groups of prompted stages. In each stage participants were asked to follow along a video of a set of example movements, either a mix of discrete gestures or freeform unprompted movements. Stages lasted 45 to 60s, while freeform stages lasted 60 to 120s.
During data collection, users donned on-and-off the device on average 3.9 times in total, see \cref{table:dataset_details}. We call these \textit{sessions} and are clearly annotated in our dataset. 
Participants performed all movements while standing or sitting on a tall stool. During each stage participants were asked to move their hand from right to left and up and down to ensure a broad range of postures.

\subsection{Stage Descriptions}
\label{sec: stage_descriptions}

The data collection protocol was designed to capture a wide range of kinematics. Each stage consistent of a particular set of instructed kinematics. Descriptions of the kinematics performed in each stage can be found in \cref{table:stage_dataset_details}.

\begin{table}[ht!]
    \scriptsize
    \centering
    \caption{\textit{Stage descriptions.} For video examples, visit \url{https://github.com/facebookresearch/emg2pose}.}
    \renewcommand{\arraystretch}{1.2}  % Add more space between rows
    
    % The following is pasted from ctrl-jupyter
    \begin{tabular}{ll}
    \toprule
    Stage & Movements \\
    \midrule
    \multirow[t]{3}{*}{FingerPinches1} & Finger pinches (4) \\
     & D-pad style thumb swipes (4) \\
     & Thumb rotations \\
    \cline{1-2}
    \multirow[t]{2}{*}{Object1} & Drink from and rotate a cup \\
     & Squeeze a soft toy \\
    \cline{1-2}
    Counting1 & Fingers counting up and down (5) \\
    \cline{1-2}
    \multirow[t]{3}{*}{Counting2} & Fingers counting up and down (5) \\
     & Wiggling the fingers \\
     & Abducting and adducting the fingers \\
    \cline{1-2}
    \multirow[t]{3}{*}{DoorknobGrab} & Mimic opening a doorknob \\
     & pull fingers into a loose fist \\
     & use index fingers as a trigger \\
    \cline{1-2}
    \multirow[t]{2}{*}{Throwing} & Mimic swinging ping pong paddle \\
     & Mimic throwing a ball \\
    \cline{1-2}
    Abduction & Series of finger abduction movements \\
    \cline{1-2}
    FingerFreeform & Freeform movements of fingers \\
    \cline{1-2}
    FingerPinches2 & Single and multiple finger pinches (multiple fingers touching the thumb simultaneously)\textbf{} (7) \\
    \cline{1-2}
    \multirow[t]{3}{*}{HandHandInteractions} & Slide fingers along the palm \\
     & Clap hands \\
     & Co-wiggle fingers \\
    \cline{1-2}
    Wiggling1 & Wiggling and spreading fingers \\
    \cline{1-2}
    \multirow[t]{2}{*}{Punch} & Pull hand torwards body while grasping fingers \\
     & Punching motion \\
    \cline{1-2}
    \multirow[t]{3}{*}{Gesture1} & Form and unform a claw \\
     & Bring fingers together in loose fist \\
     & Flick fingers individually (4) \\
    \cline{1-2}
    \multirow[t]{2}{*}{StaticHands} & Move hand from waist level to chest \\
     & Press hands together and move slowly \\
    \cline{1-2}
    FingerPinches3 & Like FingerPinches1, but with the hands occluding eachother \\
    \cline{1-2}
    Wiggling2 & Like Counting2, but with the hands occluding eachother \\
    \cline{1-2}
    Unconstrained & The hand that is not prompted to move during a particular stage \\
    \cline{1-2}
    \multirow[t]{2}{*}{Gesture2} & Extend index and pinky while curling middle fingers \\
     & Make scissor cutting motion \\
    \cline{1-2}
    \multirow[t]{3}{*}{FingerPinches2} & Index finger pinches \\
     & Middle finger pinches \\
     & D-pad style thumb swipes (4) \\
    \cline{1-2}
    \multirow[t]{2}{*}{Pointing} & Point individual fingers (5) \\
     & Snap middle finger and thumb \\
    \cline{1-2}
    Freestyle1 & Free style movements with one hand \\
    \cline{1-2}
    \multirow[t]{2}{*}{Object2} & Play with blocks \\
     & Move chess pieces \\
    \cline{1-2}
    \multirow[t]{3}{*}{Draw} & Poking \\
     & Mimic drawing \\
     & Pinching, rotating with hands both close and far \\
    \cline{1-2}
    \multirow[t]{2}{*}{Poke} & Poking \\
     & Pinch and rotate wrist \\
    \cline{1-2}
    \multirow[t]{3}{*}{Gesture3} & Extend thumb and pinky, curl middle fingers \\
     & Vulcan salute \\
     & Peace sign \\
    \cline{1-2}
    \multirow[t]{2}{*}{ThumbsSwipes} & D-pad style thumb swipes (4) \\
     & Slowly fold and unfold all fingers \\
    \cline{1-2}
    ThumbRotations & Thumb rotations \\
    \cline{1-2}
    Freestyle2 & Free style movement with both hands \\
    \cline{1-2}
    \multirow[t]{2}{*}{WristFlex} & Wrist flexion \\
     & Wrist abduction \\
    \cline{1-2}
    \end{tabular}
    
    \vspace{3mm}
    \label{table:stage_dataset_details}
\end{table}

\subsection{Dataset Limitations} 
\label{sec: dataset limitations}
% VERSION 2
While our dataset is the largest and highest fidelity open-sourced to date, it is smaller than those used in \citet{ctrl2024generic}, which may hinder generalization. While we provide high quality pose labels from motion capture using the inverse kinematics approach from \citet{han2018online}, as a camera-based method it still suffers from occlusion, hindering label quality for gestures such as fist clenching. We additionally do not track wrist movements, which are important for how we interact with the world. Alternate labelling methods, such as stretch-sensing gloves, could address these limitations, at the potential expense of lower quality labels and impaired dexterity. Finally, future datasets could include both camera and sEMG sensors, which could be combined to improve pose inference in contexts where camera-based tracking fails such as occlusion. 

\subsection{Ethical and Societal Implications}
\label{sec: societal implications}

The broader usage of sEMG and the specific development of sEMG pose estimation models may pose novel ethical and societal considerations. A highly performant emg-to-pose model running on a device placed on the wrist or forearm could store or transmit information about a person’s actions, and appropriate safeguards to encrypt and limit access to this information may be warranted. There are numerous societal benefits for the development of sEMG models for pose estimation. sEMG allows one to directly interface a person’s neuromotor intent with a computing device. This can be used to create novel device controls for the general population and can also be used to develop adaptive controllers for those who struggle to use existing computer interfaces.

\subsection{Dataset Instructions}
\label{sec: dataset_instructions}

Data is hosted on Amazon S3, and code with readme instructions as to how to reproduce experimental results are found on: \url{https://github.com/facebookresearch/emg2pose}. 

\section{Algorithm details}
\label{sec: algorithm_details}

\subsection{vemg2pose}
\textit{vemg2pose} consists of a convolutional \textit{featurizer} and an LSTM \textit{decoder}. First, the featurizer converts sEMG to features:

\begin{align}
    \mathbf{z}_t = f(\mathbf{e}_{t-R : t})    
\end{align}

where $f$ is the featurizer, $R$ is the featurizer receptive field, and $\mathbf{e}_t$ and $\mathbf{z}_t$ are vectors of sEMG and features at time $t$, respectively. Next, the decoder produces angular velocity predictions as a function of the features and the previous joint angles. Velocities are integrated to produce angular predictions:

\begin{align}
    \mathbf{s}^v_t &= \pi(\mathbf{z}_t, \mathbf{s}_{t-1}) \\
    \mathbf{s}_t &= \mathbf{s}_{t-1} + \mathbf{s}^v_t    
\end{align}

where $\pi$ is the decoder, $\mathbf{s}_t$ is the angular prediction at time $t$, and $\mathbf{s}^v_t$ is the angular velocity prediction at time $t$. For the \textit{tracking} task, the ground truth first state is provided: $\mathbf{s}_0 := \mathbf{s}^*_0$. For \textit{regression}, the ground truth state is unknown. Therefore, the decoder produces angle and angular velocity predictions ($\mathbf{s}^p$ and $\mathbf{s}^v$, respectively). The angular predictions are used for the first $P$ time steps ($250$ ms in our case), and velocities are integrated thereafter:

\begin{align}
    \mathbf{s}^p_t, \mathbf{s}^v_t &= \pi(\mathbf{z}_t, \mathbf{s}_{t-1}) \\
    \mathbf{s}_t &= \begin{cases}
        \mathbf{s}^p_{t} & \text{if } t < P, \\
        \mathbf{s}_{t-1} + \mathbf{s}^v_t & \text{if } t \geq P
    \end{cases}
\end{align}

The LSTM has two hidden layers of size 512. We scale its output by $.01$, as we find that this improves training. A \textit{Time-Depth Separable Convolution} (TDS) network is used for the featurizer, as it has been shown to be effective in the automatic speech recognition literature \citep{hannun2019sequence}. The featurizer first applies three 1D convolutions over time with 256 features, kernel widths of 11, 5, and 17, and strides of 5, 2, and 4. There are then 4 TDS blocks with channel and feature widths of 16 and kernel widths of 9, 9, 5, and 5. Overall, the featurizer reduces the sample rate to 25 Hz, and a final linear up-sampling brings them to 50 Hz. We use layer norms as described in \citep{hannun2019sequence}.

\subsection{NeuroPose}

We implement the NeuroPose U-Net architecture as described in \citet{liu2021neuropose}, with minor modifications to account for differences in recording device and joint angle targets. The encoder of the original NeuroPose has 40x temporal down-sampling achieved via a series of strides. To account for the 10x greater sample rate of our device, we double each of 3 temporal strides to yield 360x down-sampling. Similarly, we double the spatial stride of the final encoder convolution to account for the 2x spatial resolution of our device. We similarly modify the up-sampling in the decoder by the same factors and add a final linear project to achieve 20 dimensional angular predictions.

Note that the original NeuroPose uses a velocity regularization term, which we do not explore here. We find that predicting velocities rather than joint angles is sufficient to achieve smooth predictions, and precludes having to tune the weight on the velocity regularization term.

\subsection{SensingDynamics}

The original SensingDynamics was designed for a high-density sEMG device with 320 electrodes spread over 5 separate patches on the forearm and wrist \citep{simpetru2022sensing}. The architecture uses 3d convolutions over channels, patches, and time. In contrast, the sEMG-RD wrist band from \citet{ctrl2024generic} does not have separate patches and has distinct channel densities and temporal resolutions. To account for these discrepancies, we use 2d convolutions over sEMG channels and time, and modified the convolutional kernel, strides and dilations to match the effective receptive fields and strides of the original setup. 

Note that the original SensingDynamics smooths the output predictions with a moving average filter of 150 ms. We find that predicting velocities rather than joint angles is sufficient to achieve similarly smooth predictions.

\subsection{Training Setup}
\label{sec: training_setup}

For each algorithm, we performed a hyperparamter sweep over the following parameters, with each explored independently: training window length (2000-12000 samples at 2kHz, 3 different values), learning rate (.001 or .0001), gradient norm clipping (none or 1), and whether the decoder used an MLP or LSTM (for (v)emg2pose). The most performant setting was used for each algorithm, as reported in \cref{table:hyperparameters} (for emg2pose explanation see \cref{sec: vel-vs-pos}). A learning rate of 0.001 was universally optimal. To improve generalization across device placements, we use \textit{rotation augmentation}, wherein we spatially rotate the sEMG channels by $1$, $0$, or $-1$ (uniformly sampled). Augmentation is only applied during training.

\begin{table}[ht!]
    \footnotesize
    \centering
    \caption{\textit{Algorithm comparison.} We extend the window lengths to account for receptive fields. 
    }
    \renewcommand{\arraystretch}{1.5}  % Add more space between rows
    
    % Paste from ctrl-jupyter
    \begin{tabular}{llllll}
    \toprule
     &  Baseline & Predictions & Network & Grad clip & Window \\
    \midrule
    \multirow[t]{2}{*}{Tracking} & emg2pose & Angles & TDS + MLP & 1 & 5790 \\
     & vemg2pose & Velocities & TDS + LSTM & 1 & 11790 \\
    \cline{1-6}
    \multirow[t]{3}{*}{Regression} & emg2pose & Angles & TDS + MLP & 1 & 5790 \\
     & vemg2pose & Velocities & TDS + LSTM &  0 & 11790 \\
     & NeuroPose & Angles & U-Net &  0 & 4000 \\
     & SensingDynamics & Angles & 2d Conv + MLP & 0 & 10167 \\
    \cline{1-6}
    \end{tabular}

    \vspace{3mm}
    \label{table:hyperparameters}
\end{table}

\subsection{Online vemg2pose}
\label{sec: online_vemg2pose}

To enable online deployment of vemg2pose it must be setup to handle sEMG data being received sequentially in discrete packets of variable temporal lengths. As such, we created a variant of vemg2pose that uses buffers to append the current packet of data to the previous ones. In addition to storing all received data in a buffer, we additionally keep track of which data have been processed already and which have not (this is a function of the network receptive field and the stride), so as not to make duplicate predictions. For \cref{fig: EMG2Pose online}, we trained a \textit{vemg2pose, tracking} model with this internal variant. This internal variant achieved joint angular errors almost identical to those reported in \cref{table:tracking_results}, as expected. We do not open-source this setup, as it is only useful with access to the sEMG-RD band for online testing. 

\subsection{Hand mesh visualizations of prediction trajectories}
\label{sec: mesh_visualization}

We generated the articulated hand meshes representing prediction trajectories (as depicted in \cref{fig: online rollout} and \cref{sec: track_ex}) from sequences of joint angles using the forward kinematic and a mesh-skinning algorithms provided by UmeTrack \citep{han2022umetrack}. We use the generic, default hand model provided by UmeTrack. To generate the figures, we render the meshes using the Plotly and Plotly-Kaleido visualization packages \citep{plotly}.

\subsection{Statistical Analysis}
\label{sec: statistical_analysis}
The Wilcoxon statistical analyses reported in \cref{table:regression_results} were performed on data aggregated across time for each user. That is, metrics were computed at each temporal sample, then averaged across time for each user within each experimental condition. Statistics aggregated within each user and condition are similarly used to construct distributions for all other plots and tables.

\section{Detailed Analyses}
\label{sec: detailed_analysis}

\subsection{Analysis of Stages that are Challenging for Vision-Based Systems}
\label{sec: challenging_stages_supp}

We compared the same stages with and without occlusion, and found that occlusion did not negatively impact model performance, as expected (\cref{fig: challenging_stages}, left). Each subject performed the CountingWiggling and FingerPinches stages under two conditions: with the hands in front them - such that they would be visible to a headset based CV tracking system - and with the hands very close to or very far away from the body - such that they would be occluded.

We also compared stages with hand-object interactions, hand-hand interactions, and no interactions (\cref{fig: challenging_stages}, right). Hand-object interactions consisted of the Object1 and Object2 stages, in which participants interacted with a cup, a soft toy, blocks, and chess pieces. Hand-hand interactions (HandHandInteractions stage) consisted of sliding the fingers across the opposite palm, clapping the hands together, and wiggling the fingers such that the fingertips of opposite hands tap against one another. These interaction types are known to be challenging for vision-based systems. Nonetheless, performance for these stages was comparable or superior to performance in stages without any interactions. Note, however, that the behavioral distributions are different across these stages, which makes direct comparison of metrics challenging.

\subsection{Velocity vs. Positional Predictions}
\label{sec: vel-vs-pos}

\begin{figure}[h!]
  \centering
  \includegraphics[width=1\textwidth]{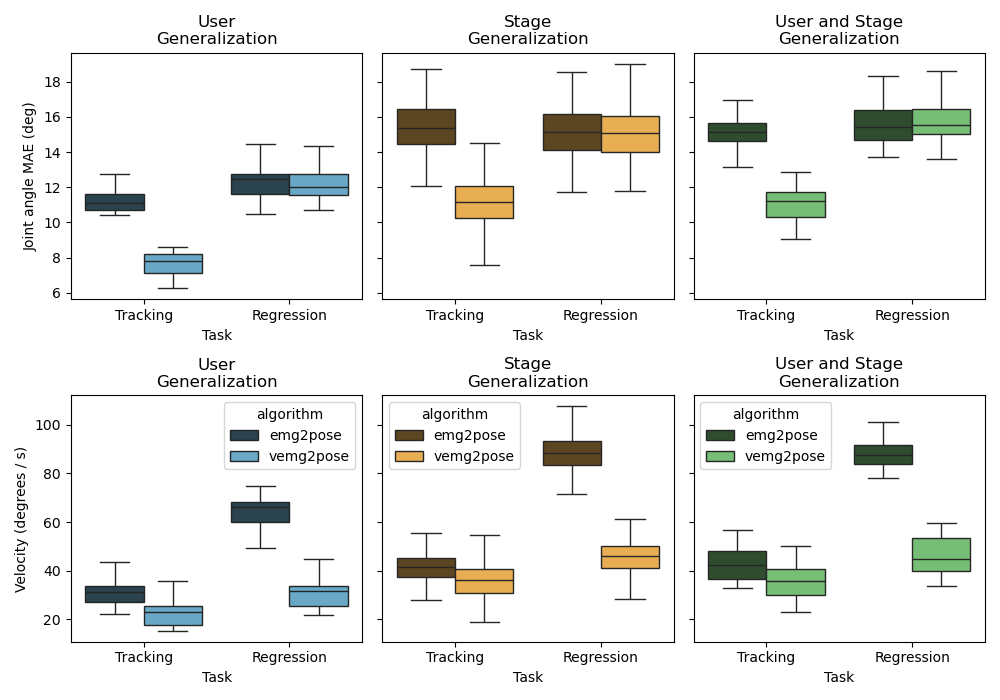}
  \caption{\textit{vemg2pose} vs. \textit{emg2pose} for tracking and regression tasks. Distributions are over users. Box plots take the same format as \cref{fig: results stage breakdown}.}
  \label{fig: emg2pose vs vemg2pose}
\end{figure}

We compared \textit{vemg2pose} to \textit{emg2pose}, an otherwise identical algorithm that directly predicts joint angles rather than joint angular velocities (\cref{fig: emg2pose vs vemg2pose}). emg2pose has similar joint angular error in the regression task, but much worse performance on the tracking task. This is likely because vemg2pose is initialized to the ground truth initial state, whereas emg2pose is merely conditioned on the ground truth initial state. For both tracking and regression tasks, vemg2pose has lower overall velocity than emg2pose, suggesting that operating in velocity space encourages smoother predictions.

\subsection{LSTM vs Transformer Decoders}
\label{sec: lstm-vs-trans}

% REGRESSION
\begin{table}[ht]
    \small
    \centering
    \caption{\textit{Regression ablation} test set results. Mean and standard deviation are reported across users.}
    \renewcommand{\arraystretch}{1.2}  % Add more space between rows
    
    % The following is pasted from ctrl-jupyter
    \begin{tabular}{lcccccc}
    \toprule
     Test Set &  Ablation & Angular Error ($\degree$) & Landmark Distance ($mm$) \\
    \midrule
    \multirow[t]{2}{*}{User}
     & vemg2pose-tran & 12.5 $\pm$ 1.3 & 16.3 $\pm$ 1.8 \\
     & vemg2pose-lstm & 12.2 $\pm$ 1.3 & 15.8 $\pm$ 1.9 \\
    \cline{1-4}
    \multirow[t]{2}{*}{Stage} 
     & vemg2pose-tran & 16.1 $\pm$ 1.6 & 21.7 $\pm$ 2.1 \\
     & vemg2pose-lstm & 15.2 $\pm$ 1.6 & 20.4 $\pm$ 2.2 \\
    \cline{1-4}
    \multirow[t]{2}{*}{User, Stage} 
     & vemg2pose-tran & 16.2 $\pm$ 1.3 & 22.3 $\pm$ 1.8 \\
     & vemg2pose-lstm & 15.8 $\pm$ 1.4 & 21.6 $\pm$ 2.0 \\
    \cline{1-4}
    \end{tabular}
    \label{table:regression_lstm_vs_transformers_results}
\end{table}

We ablated over decoder architectures, specifically LSTMs and transformers as the two most widely adopted models for sequence modelling. For the transformer, we explored the widely adopted transformer encoder BERT setup \citep{kenton2019bert}. We swept over the \textit{number of layers} (2, 4, 6) and \textit{number of heads} (2, 4, 8) reporting the best for both \textit{regression} and \textit{tracking} tasks in \cref{table:regression_lstm_vs_transformers_results,table:tracking_lstm_vs_transformers_results}. In order to fit into memory (Amazon EC2 \texttt{g4dn.metal} instances which have 8x NVIDIA T4 GPUs) we had to halve the feature dimensionality of the transformer decoder. In general, the transformer performs similarly or slightly worse than the LSTM.

% TRACKING
\begin{table}[ht]
    \small
    \centering
    \caption{\textit{Tracking ablation} test set results. Mean and standard deviation are reported across users.}
    \renewcommand{\arraystretch}{1.2}  % Add more space between rows
    
    % The following is pasted from ctrl-jupyter
    \begin{tabular}{lccc}
    \toprule
     Test Set & Ablation & Angular Error ($\degree$) & Landmark Distance ($mm$) \\
    \midrule
    \multirow[t]{2}{*}{User}
     & vemg2pose-tran & 8.0 $\pm$ 1.0 & 10.7 $\pm$ 1.6 \\
     & vemg2pose-lstm & 7.7 $\pm$ 1.0 & 10.3 $\pm$ 1.5 \\
    \cline{1-4}
    \multirow[t]{2}{*}{Stage} 
     & vemg2pose-tran & 11.7 $\pm$ 1.4 & 15.9 $\pm$ 1.9 \\
     & vemg2pose-lstm & 11.2 $\pm$ 1.4 & 15.2 $\pm$ 1.9 \\
    \cline{1-4}
    \multirow[t]{2}{*}{User, Stage} 
     & vemg2pose-tran & 11.4 $\pm$ 1.1 & 16.0 $\pm$ 1.5 \\
     & vemg2pose-lstm & 11.0 $\pm$ 1.0 & 15.4 $\pm$ 1.4 \\
    \cline{1-4}
    \end{tabular}

    \vspace{-3mm}
    \label{table:tracking_lstm_vs_transformers_results}
\end{table}

\subsection{Performance Decomposition across Fingers and Joints}

\begin{figure}[h!]
  \centering
  \includegraphics[width=1\textwidth]{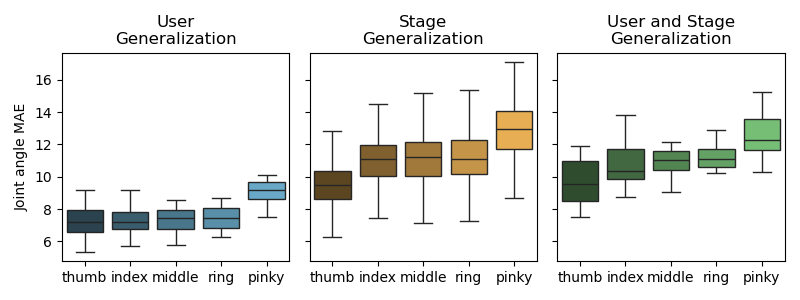}
  \caption{\textit{Performance decomposition per finger:} for tracking task, vemg2pose. Error per finger is measured by averaging the errors of the joints associated with each finger. Distributions are over users. Box plots take the same format as \cref{fig: results stage breakdown}.}
  \label{fig: finger performance}
\end{figure}
    
\begin{figure}[h!]
  \centering
  \includegraphics[width=1\textwidth]{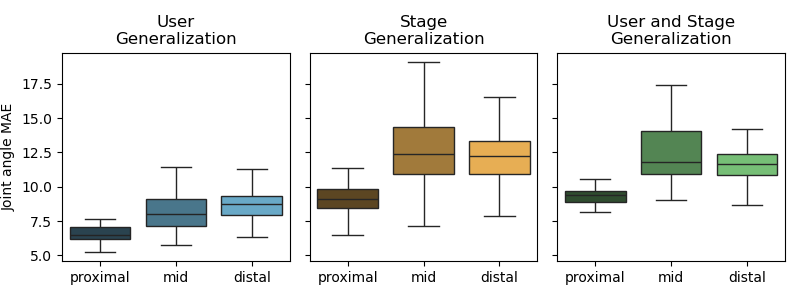}
  \caption{\textit{Performance decomposition across joint groups:} for tracking task, vemg2pose. Performance broken down by joint according to their proximal-distal location. \textit{Proximal} is CMC for the thumb and MCP for other fingers; \textit{Mid} is MCP for thumb and PIP for other fingers; and \textit{Distal} is IP for thumb and DIP for other fingers. Box plots take the same format as \cref{fig: results stage breakdown}.}
  \label{fig: joint performance}
\end{figure}

We decompose \textit{vemg2pose} tracking performance across fingers (\cref{fig: finger performance}), and proximal, mid, and distal joint groups (\cref{fig: joint performance}). For the latter, proximal, mid, and distal joints are grouped according to their distance from the palm. See \cref{fig: joint performance} for further details. Reconstruction performance varies across fingers and joint groups. Thumb and pinky fingers are consistently best and worst performers, and proximal joints are more easily predicted than distal joints.

\newpage
\subsection{Tracking Trajectory Examples}
\label{sec: track_ex}

\begin{figure}[h!]
  \includegraphics[width=1\textwidth]{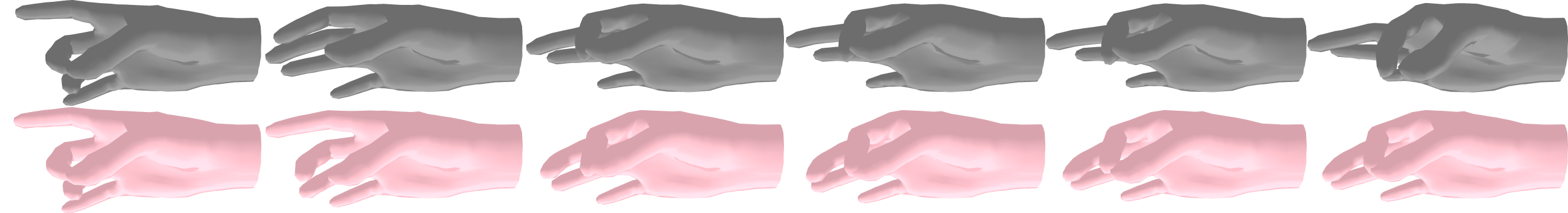}
  \caption{\textit{Held-Out User, Stage} tracking, top 15\% stage (Gesture2), median user. Top: motion capture; bottom: vemg2pose, tracking predictions. Clips unroll evenly left-to-right over a $2$ seconds.}
  \label{fig: v1}
\end{figure}

\begin{figure}[h!]
  \includegraphics[width=1\textwidth]{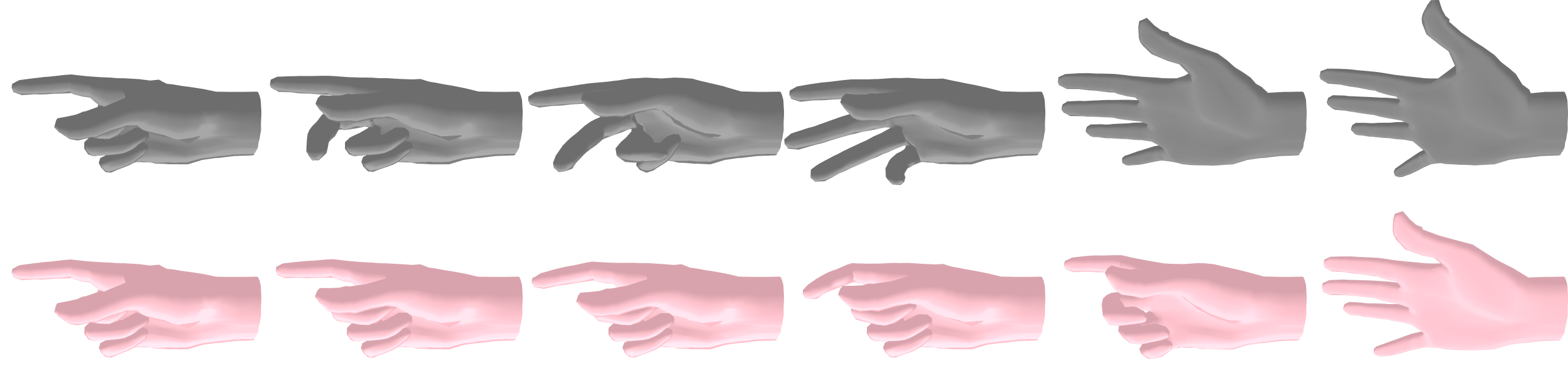}
  \caption{\textit{Held-Out User, Stage} tracking, bottom 15\% percentile stage (Counting1), median user. Top: motion capture; bottom: vemg2pose, tracking predictions. Clips unroll evenly left-to-right over a $2$ seconds.}
  \label{fig: v2}
\end{figure}

\begin{figure}[h!]
  \includegraphics[width=1\textwidth]{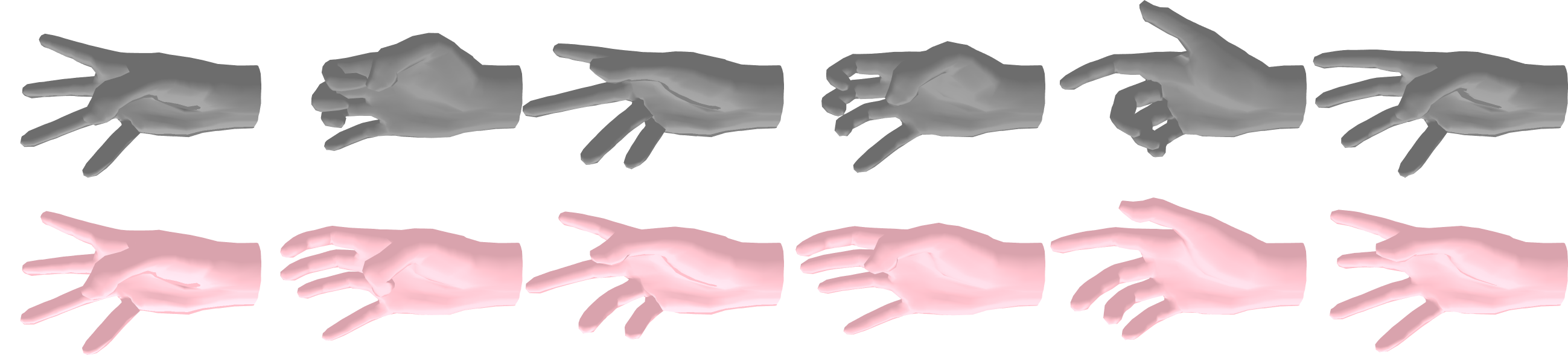}
  \caption{\textit{Held-Out User, Stage} tracking, median stage (Counting2), top 15\% percentile user. Top: motion capture; bottom: vemg2pose, tracking predictions. Clips unroll evenly left-to-right over a $2$ seconds.}
  \label{fig: v3}
\end{figure}

\begin{figure}[h!]
  \includegraphics[width=1\textwidth]{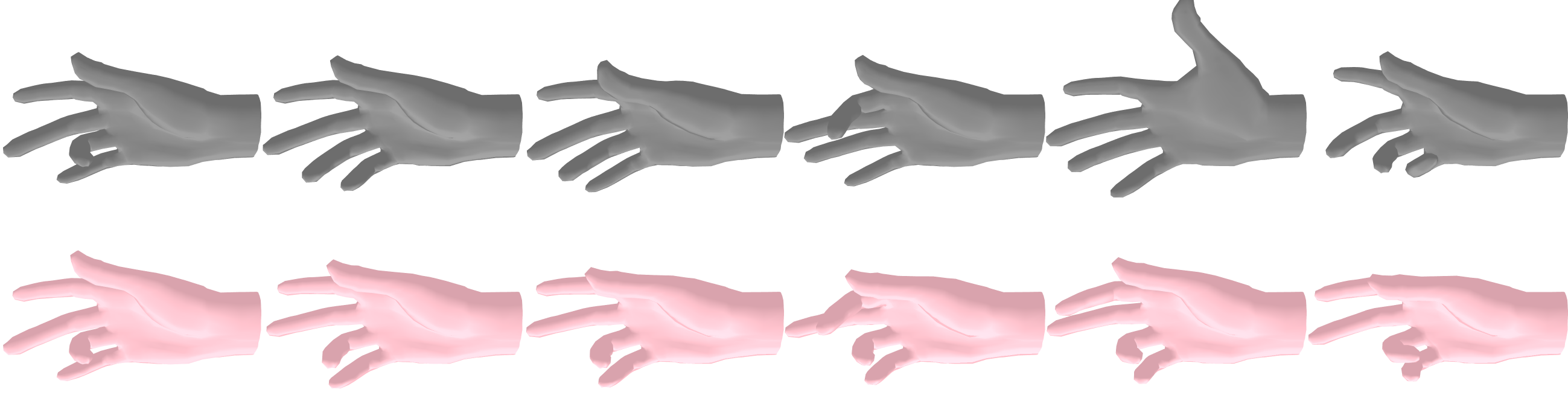}
  \caption{\textit{Held-Out User, Stage} tracking, median stage (Wiggling2), bottom 15\% percentile user.  Top: motion capture; bottom: vemg2pose, tracking predictions. Clips unroll evenly left-to-right over a $2$ seconds.}
  \label{fig: v4}
\end{figure}

We provide representative \textit{vemg2pose, tracking} prediction trajectories for the $15$\%, $50$\% and $85$\% user and stage percentiles for the held-out \textit{user, stage} scenario described in \cref{sec: held-out settings}. Performance varies considerably across held-out users and stages, as seen in \cref{fig: v1,fig: v2,fig: v3,fig: v4}. We note that there may exist large variance \textit{within} stages, for which these kinematic plots do not reflect. Minimizing these variances will be of great value. For video examples, visit \url{https://github.com/facebookresearch/emg2pose}. 

\newpage

% \section{Ethical and Societal Implications}
% \label{sec: societal implications}
% Our dataset contains de-identified time series of sEMG and hand joint angles, without specific hand size information. sEMG data and hand movement data are not considered biometric data in any legal jurisdiction and do not give the dataset user the ability to identify individuals who participated in the research study, and as a result do not pose a privacy threat to those who participated in the study. Nonetheless, the broader usage of sEMG and the specific development of sEMG pose estimation models pose novel ethical and societal considerations. A well performing EMG2Pose model running on a device placed on the wrist or forearm could store or transmit information about a person’s actions, and appropriate safeguards must be put in place to encrypt and limit access to this information. sEMG sensing reads out biological signals of potential clinical significance and so must be appropriately protected. 

% There are also societal benefits for the development of sEMG models for pose estimation. sEMG allows one to directly interface with a persons’ neuromotor intent, giving the ability to directly interface. This can be used to develop adaptive controllers for those who struggle to use existing computer interfaces. 

\end{document}